\documentclass[journal]{IEEEtran}
\usepackage{amsmath,amsfonts}
\usepackage{algorithmic}
\usepackage{algorithm}
\usepackage{array}
\usepackage[caption=false,font=normalsize,labelfont=sf,textfont=sf]{subfig}
\usepackage{textcomp}
\usepackage{stfloats}
\usepackage{verbatim}
\usepackage{cite}
\usepackage{diagbox}
\usepackage[utf8]{inputenc} % allow utf-8 input
\usepackage[T1]{fontenc}    % use 8-bit T1 fonts
\usepackage{hyperref}       % hyperlinks
\usepackage{url}            % simple URL typesetting
\usepackage{booktabs}       % professional-quality tables
\usepackage{amsfonts}       % blackboard math symbols
\usepackage{nicefrac}       % compact symbols for 1/2, etc.
\usepackage{microtype}      % microtypography
\usepackage{xcolor}         % colors
\usepackage{graphicx}
\usepackage{multirow}
\usepackage{multicol}
\usepackage{wrapfig}
\usepackage{enumitem}
\usepackage{amsmath}
\usepackage{float}
\usepackage{makecell}
\usepackage{flushend}
\setlist[itemize]{noitemsep, topsep=0pt}
\usepackage{pifont}% http://ctan.org/pkg/pifont
\newcommand{\cmark}{\ding{51}}%
\newcommand{\xmark}{\ding{55}}%

\usepackage{color}
\definecolor{revred}{RGB}{180,40,40}
\definecolor{revgreen}{RGB}{30,120,70}
\definecolor{revblue}{RGB}{35,90,170}

\begin{document}

\title{BVI-RLV: A Fully Registered Dataset for Low-Light Video Enhancement}

\author{%
  Ruirui~Lin,  Guoxi~Huang,  {Joanne~Lin, Qi~Sun, Alexandra~Malyugina, David~R.~Bull,~\IEEEmembership{Fellow,~IEEE,} and Nantheera~Anantrasirichai,~\IEEEmembership{Member,~IEEE,}} 

  % <-this % stops a space
\thanks{All authors are with Visual Information Laboratory, Bristol Vision Institute (BVI), University of Bristol, Bristol, UK BS1 5DD. This work was supported by the UKRI MyWorld Strength in Places Programme (SIPF00006/1) and  EPSRC NIA (UKRI241) }% <-this % stops a space
}

\markboth{Journal of \LaTeX\ Class Files,~Vol.~14, No.~8, August~2021}%
{Shell \MakeLowercase{\textit{et al.}}: A Sample Article Using IEEEtran.cls for IEEE Journals}


\maketitle

\begin{abstract}
Low-light videos often exhibit spatiotemporally incoherent noise, compromising visibility and degrading performance in computer vision applications. A major challenge for enhancing such content using deep learning lies in the scarcity of pixel-aligned, high-quality training data.
We introduce BVI-RLV, a fully registered low-light video dataset comprising over 30k paired frames from 40 diverse scenes under two low-light conditions, each aligned with normal-light ground truth. Unlike existing datasets that rely on neutral density (ND) filters or suffer from misalignment issues, BVI-RLV achieves sub-pixel registration for 99.24\% of data at full HD resolution across dynamic motion scenarios using a motorized dolly and image-based refinement. The dataset covers a wide range of motion types and realistic temporal noise. We also provide baseline implementations using four representative architectures: Convolutional Neural Network (CNN), Transformer, State Space Model (Mamba), and Diffusion Model (DM). Experiments demonstrate that registration is crucial for supervised learning, yielding up to 5.85~dB PSNR improvement compared to unregistered training. Models trained on BVI-RLV outperform those trained on existing datasets in cross-dataset evaluations, achieving superior performance even in real-world outdoor scenes. Our dataset is publicly available at \url{https://doi.org/10.21227/mzny-8c77}.
\end{abstract}


\section{Introduction}
\IEEEPARstart{C}{apturing} videos in low-light conditions is necessary for various applications, including natural history filmmaking, biology, zoology, robotics, surveillance, and security. However, low-light video footage is problematic due to the interactions between the aperture, shutter speed, and ISO parameters,  resulting in various types of distortions. This arises from the inverse correlation of increased photon noise with decreasing light intensity, where increased sensor gain amplifies noise. Secondary characteristics such as white balance and color effects are also affected and require correction. While these distortions reduce perceived visual quality, the associated artifacts pose significant challenges to machine learning tasks such as classification and detection. %This is especially true considering the effect of adversarial examples on some deep learning networks.

The performance of low-light \textit{image} enhancement algorithms has improved significantly in recent years, primarily through the application of deep learning methods~\cite{SGF_2024,Jiang:Low:2023,LIE_2024}. Such techniques, however, cannot be applied directly to low-light \textit{video} content as their application on a frame-by-frame basis causes temporal inconsistencies. 
Video enhancement optimization is fundamentally limited by the availability of datasets with accurate spatial-temporal alignment. Unfortunately, low-light video enhancement (LLVE) is ill-posed because obtaining pixel-aligned ground truth with accurate brightness and motion is highly challenging.

% \IEEEpubidadjcol %clear out second column to show the IEEEpubid mark
 
\begin{figure}[t!]
  \centering
    \includegraphics[trim={3mm 1cm  15mm 20mm}, clip, width=0.95\columnwidth]{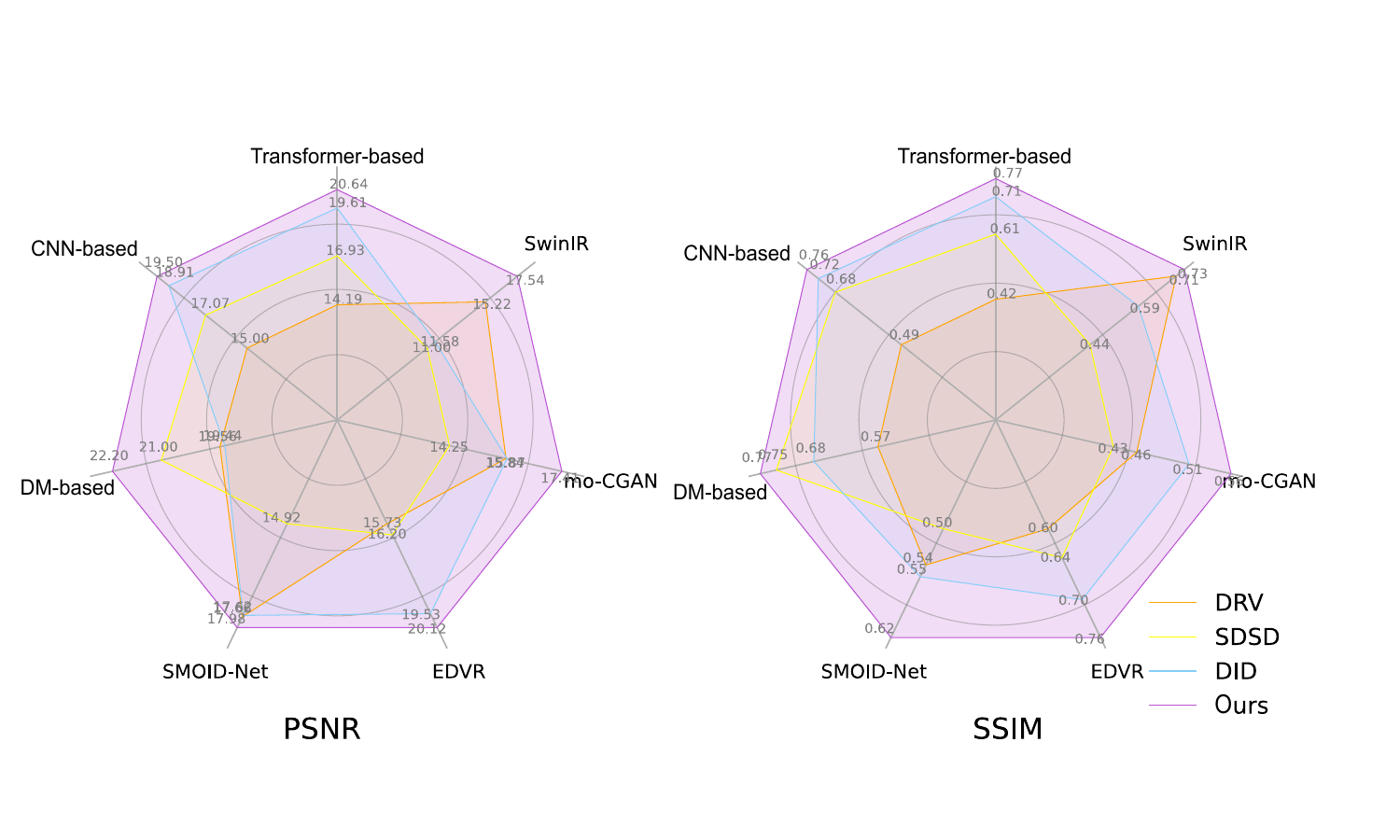}
    %\vspace{-0.3cm}
    \caption{Models trained on ours (BVI-RLV) generate better results compared to the other three LLVE datasets (i.e., DRV~\cite{Chen_2019_ICCV}, SDSD~\cite{wang2021sdsd}, DID~\cite{Fu:dancing:2023}). PSNR and SSIM are averaged (unweighted) across all four test sets.}
    \label{fig: radar_psnr}
   %\vspace{-4mm}
\end{figure}

% The scarcity of aligned and accurate ground truth data continues to hinder progress in low-light video research. Hence, many studies rely on synthesized datasets but may fail to replicate real distortions. Although synthetic image samples have been explored~\cite{9008378}, such distortions are typically oversimplified. Real low-light conditions involve multiple degradation types, and simulating them from static images cannot capture temporal noise. Motion and focus blur further complicate video characteristics, making stop-motion approaches unsuitable. Existing LLVE datasets either lack motion diversity or suffer from frame misalignment, both of which hinder supervised learning. Self-supervised approaches~\cite{Dewil:Self:2021} offer partial contrast enhancement but are ill-suited for color correction, as low-light videos lack accurate color information -- an issue confirmed by~\cite{Liang:self:2022}. Alternatively, unpaired learning strategies can be used~\cite{anantrasirichai:Contextual:2021,CycleRetinex_2024,NVEU_2023}, though they are computationally expensive and memory-intensive due to dual-network designs and adversarial objectives.

The scarcity of aligned and accurate ground-truth data remains a major bottleneck for low-light video research. Many studies therefore rely on synthetic datasets, but simplified image-based degradations ~\cite{9008378} cannot fully capture real video noise, motion blur, focus variation, or temporal artifacts. Existing LLVE datasets either have limited motion diversity or suffer from frame misalignment. This issue is particularly detrimental to supervised learning, as pixel-wise losses (e.g., $\ell_1$) are highly sensitive to small spatial offsets. Misalignment may introduce blur and ghosting artifacts during training and bias quantitative evaluation. %; in Section~\ref{sssec:fullyregister_exp}, accurate registration improves PSNR by up to 5.85~dB over unregistered training.
 Self-supervised approaches~\cite{Dewil:Self:2021} offer partial contrast enhancement but remain limited for color correction, as low-light videos lack reliable color information to guide restoration~\cite{Liang:self:2022}; unpaired strategies~\cite{anantrasirichai:Contextual:2021,CycleRetinex_2024,NVEU_2023} can alleviate the need for paired data but are computationally expensive due to dual-network designs and adversarial objectives.

To address these challenges, we present BVI-RLV (\underline{R}egistered 
\underline{L}ow-light \underline{V}ideos), a fully registered dataset of real 
low-light videos with normal-light ground truth. BVI-RLV combines three 
essential characteristics for effective supervised learning: (1) 99.24\% of 
data pairs achieve sub-pixel registration accuracy, verified using 
cross-correlation with an alignment error of $0.256 \pm 0.451$ pixels, 
further supported by corner detection ($0.619 \pm 0.610$ pixels);
(2) authentic noise and motion blur from real captures under two controlled illumination levels (10\% and 20\% of normal lighting), with 100\% lighting serving as ground truth; and (3) diverse motion patterns (linear, nonlinear, rotational, and angled), offering the highest content diversity among existing datasets. BVI-RLV comprises 40 varied indoor scenes and over 30k paired frames, all registered at full HD resolution. Dynamic sequences were captured using a programmable motorized dolly to ensure repeatable motion profiles. Indoor acquisition enables precise registration, which is infeasible in uncontrolled outdoor environments due to lighting and mechanical constraints. Models trained on BVI-RLV outperform those trained on existing datasets (as shown in Fig.~\ref{fig: radar_psnr}), confirming the critical importance of accurate registration for supervised learning.

% Given the limited availability of high-quality datasets, few low-light video enhancement methods exist. We therefore implement four representative baseline architectures to demonstrate the dataset’s benchmarking value and reproducibility. Furthermore, we show that by carefully designing scenes with essential content and motion, models trained on our indoor dataset can enhance outdoor scenes to appear more natural and subjectively superior to those produced by models trained on less curated outdoor datasets. To conclude, our contributions can be summarized as follows:
Given the limited availability of high-quality datasets, few LLVE methods exist. We therefore implement four representative baseline architectures to demonstrate the benchmarking value and reproducibility of BVI-RLV. We further show that carefully designed indoor scenes with diverse content and motion enable models trained on BVI-RLV to generalize well to outdoor scenes. Our contributions are summarized as follows:

\begin{enumerate}
    \item We introduce a novel, fully registered low-light video dataset with ground truth references, comprising 40 diverse scenes, various motion patterns, and over 30k paired frames with pixel-wise alignment at full HD resolution.
    \item We analyze the performance of the proposed BVI-RLV dataset, demonstrating the importance of full registration for low-light video enhancement, and benchmark it against three widely used existing datasets.
    \item We conduct an extensive evaluation of six state-of-the-art enhancement methods and four standardized baselines based on distinct architectures (CNN, Transformer, Mamba, and DM), ensuring transparent and consistent training and testing across multiple datasets.
\end{enumerate}

%%\vspace{-0.1cm}
\section{Related work}
%%\vspace{-0.1cm}
\subsection{Existing low-light datasets}

Several `\textit{image}'-based datasets exist that can be used to support denoising and related inverse problems. These include the UCID-v2 corpus \cite{5651245}, the Darmstadt Noise Dataset \cite{8099777}, and the Smartphone Image Denoising Database (SIDD)~\cite{abdelhamed2018high}. A few video-based datasets also exist, primarily captured in static scenes to enable ground
truth generation. This includes BVI-LOWLIGHT~\cite{Malyugina:topological:2023}, which offers true ISO noise in videos captured at 30 frames per scene, and DRV (also referred to as SMID) \cite{Chen_2019_ICCV}, which captures 60-100 frames per scene. Notably, DRV also includes dynamic videos without ground truth, similar to the BDD100K dataset~\cite{Yu_2020_CVPR}. The Dancing-under-the-stars dataset (DUS) \cite{Monakhova:DUS:2022} provides 26 static scenes, each with one clean image paired with noisy bursts, and 42 unpaired noisy dynamic videos for testing. The raw videos were captured with a high-end Canon LI3030SAI sensor, which provides an additional near-infrared channel that they include to perform enhancement.
Table \ref{tab:existingdata} summarizes existing low-light datasets with dynamic content. Among them, only five datasets contain motion and provide ground truth references:

\begin{enumerate}[leftmargin=0pt,topsep=0pt,itemsep=-0.75ex,partopsep=1ex,parsep=1ex,itemindent=1.2em]
\item \textbf{SMOID} \cite{Jiang:learn:2019} consists of 179 pairs of street view videos, each with 200 frames in a GBRB Bayer pattern, captured simultaneously using a beam splitter and an ND filter. Only overlapping regions were used. Unfortunately, this dataset is not publicly available.
\item  \textbf{RViDeNet} \cite{9156826} was captured at five ISO levels from 1600 to 25600. Each video contains seven Bayer frames captured via stop motion, resulting in no motion blur, large inter-frame displacements, and unnatural motion. The clean frame, obtained by averaging multiple noisy frames, remains dark. Therefore, this dataset is unsuitable for brightness enhancement and color correction in low-light video enhancement.
\item  \textbf{LLRVD} \cite{Fu_2023} contains 210 video pairs in RGBG Bayer format, created for low-light video denoising by capturing dynamic scenes on a high-quality monitor. Ground truth and low-light pairs were generated with long and short exposures, respectively, but the dataset is not publicly available.

\item  \textbf{SDSD} \cite{wang2021sdsd} has 150 video pairs captured using a motorized dolly, but some frames are misaligned, making them unsuitable for supervised learning or objective evaluation. Our experiments confirmed this issue. Despite claims of using an ND filter and adjusted ISO settings, there are inconsistencies in lighting and shadows, and several reference videos are overexposed~\cite{Fu:dancing:2023}. %Examples are shown in Fig. S1 in our supplemental material (SM) and \cite{Fu:dancing:2023}.
\item  \textbf{DID} \cite{Fu:dancing:2023} has 413 video pairs, captured using five different cameras. Dynamic scenes were created with an electric gimbal, resulting in only one type of motion: panning. Video brightness was adjusted with an ND filter. Some pairs are misaligned, and some videos have dirt on the lens.
\end{enumerate}

Overall, existing datasets remain limited for developing reliable low-light video enhancement methods. Most are small in scale, lack motion diversity, exhibit unrealistic distortions, or miss ground truth for moving objects, reducing their value for training and validation. Moreover, most datasets with ground truth rely on ND filters, which often introduce color shifts, optical aberrations, reflections, and reduced sharpness, leading to degraded and inconsistent image quality.

\begin{table}[!t]
%\vspace{-1mm}
\centering
\caption{Comparison of low-light video datasets with dynamic content. Dyn. = dynamic content; Reg. Prec. = registration precision; D$^*$ = stop-motion; Avail. = public availability.}
\label{tab:existingdata}
\renewcommand{\arraystretch}{1.08}
\setlength{\tabcolsep}{2.0pt}
\resizebox{\columnwidth}{!}{
\begin{tabular}{@{}l|cccccccc@{}}
\toprule
\textbf{Name} & \textbf{Light} & \textbf{Dyn.} & \textbf{GT} & \textbf{Reg. Prec.} & \textbf{Mot. type} & \textbf{Frames} & \textbf{Res.} & \textbf{Avail.} \\
\midrule
DRV & Real & \cmark & \xmark & n/a & diverse (no GT) & 2.4k & 3672p & \cmark \\
SMOID & ND & \cmark & \cmark & frame-level & 1 type (pan) & 35.8k & 1000p & \xmark \\
RViDeNet & Real & D$^*$ & \cmark & frame-level & discont. & 385 & 1080p & \cmark \\
LLRVD & Real & \cmark & \cmark & mostly aligned & screen-replay & -- & 6336p & \xmark \\
SDSD & ND & \cmark & \cmark & partial & 1 type (linear) & 37.5k & 1080p & \cmark \\
DUS & Real & \cmark & \xmark & n/a & diverse (no GT) & 15k & 1280p & \cmark \\
DID & ND & \cmark & \cmark & partial & 1 type (pan) & 41.0k & 1440p & \cmark \\
\midrule
\textbf{BVI-RLV} & Real & \cmark & \cmark & \textbf{sub-pixel} & \textbf{4 types} & 31.8k & 1080p & \cmark \\
\bottomrule
\end{tabular}}
%\vspace{-4mm}
\end{table}

\subsection{Existing low-light video enhancement methods}
With recent advances in deep learning technologies, `\textit{image}' enhancement has made significant progress~\cite{Jiang:Low:2023,SGF_2024,Ma_2023}. In contrast, learning-based video enhancement techniques are still in their early stages~\cite{Li:Low:2022}, primarily due to more unknown parameters, limitations in datasets, computational complexity, and high memory requirements. 
Multiple input frames are typically used. SMOID \cite{Jiang:learn:2019} modified UNet to handle multiple frames, but this method may not effectively handle fast motion. Therefore, several methods align the feature maps of neighboring frames with those of the current frame~\cite{wang2021sdsd}. Earlier work exploited recurrent neural networks~\cite{Wang:enhancing:2019}, while later, Deformable Convolution Networks~\cite{Dai:Deformable:2017} became more commonly employed for this purpose. An iterative process for re-weighting features is used to further minimize merging errors~\cite{zhou2021rta}. Liu et al.~\cite{Liu:low:2023} proposed using synthetic events from video to adaptively combine multiple frames according to the speeds of moving objects. Swin Transformer is used in \cite{Lin:ICIP:2024}. Fu et al.~\cite{Fu:dancing:2023}, who created the DID dataset, also proposed a light-adjustable network based on Retinex theory. Unfortunately, their code is not publicly available. SDSD-Net~\cite{wang2021sdsd} separates noise and illumination to generate the output. The DM proposed for images \cite{Jiang:Low:2023} was adapted to videos in  \cite{Lin:CVMP:2024}. Some methods require Bayer raw video inputs, which are unsuitable for consumer cameras~\cite{Chen_2019_ICCV, Jiang:learn:2019, triantafyllidou2020low}. % Due to the limited availability of paired datasets, there are also attempts to use unpaired training strategies, such as CycleGAN \cite{anantrasirichai:Contextual:2021}. This architecture is also employed to map low-light RAW to RGB format \cite{triantafyllidou2020low}.

% ------------------------------------------------
\section{Proposed low-light video dataset}
\label{sec:newdataset}

\subsection{Acquisition settings}

\subsubsection{Controlled environment} 
\begin{figure}[t!]
    \centering
    \includegraphics[width=\columnwidth]{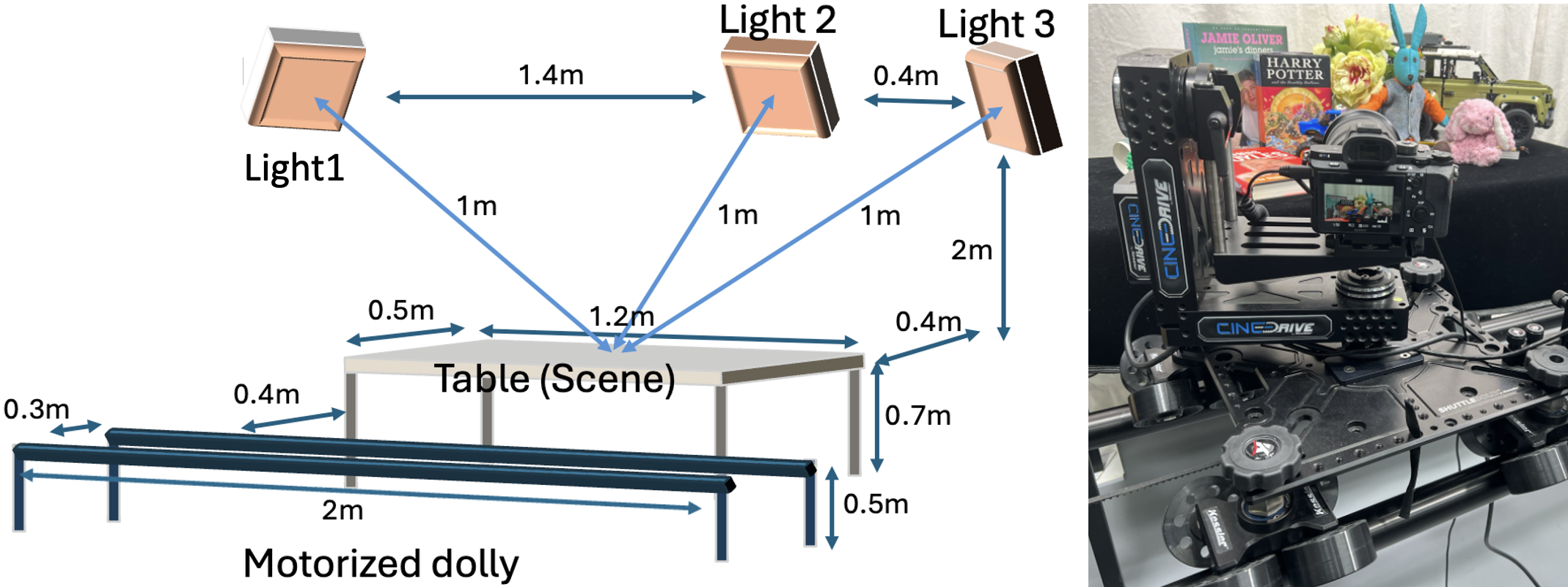}%\vspace{-3mm}
    \caption{(Left) Diagram showing the geometric position between the moving scene and the light source. (Right) Scene setting showing the camera in ‘angle’ position, mounted on CineDrive system.}
   \label{fig:viewgeo}%\vspace{-4mm}
\end{figure}

Dataset acquisition was conducted in a dedicated studio, allowing precise control of lighting and color temperature. Four Cineo MavX lights (up to 8,000 lumens each) were used, set to a color temperature of 6,500~K. The video pairs were recorded under normal lighting (100\%) and two low-light conditions (10\% and 20\% of normal intensity), calibrated using a Zero 88 FLX S24 light controller. These low-light levels correspond to approximately 3-5~lux and 10-14~lux, respectively. The geometric configuration between the moving scene and light source is illustrated in Fig.~\ref{fig:viewgeo}.

\subsubsection{Camera settings} We used a consumer camera to maximize accessibility and practical relevance. Consumer cameras output RGB-format videos; only \textit{some} high-end professional cameras support true raw video, which should not be confused with the `\textit{image}' raw format accessible on most consumer models. For readers interested in noise and the \textit{imaging} pipeline, we refer to our low-light image dataset, BVI-LOWLIGHT~\cite{Malyugina:topological:2023}, captured with the same Sony Alpha~7SII and FE~16-35mm~F2.8~GM lens. To ensure high-quality data, we minimized image signal processing (ISP) by recording with the lowest compression available on this camera -- XAVC~S format (H.264 codec, 8-bit depth, full HD 1920$\times$1080 resolution). ISO was fixed at 160 for normal light and 800 for low-light, following the manufacturer’s recommendations; higher ISO was avoided to prevent in-camera denoising. White balance was matched to the light source (C.Temp./Filter mode). Videos were captured at 25 fps with a 1/50 shutter speed (standard 180-degree shutter angle), and the aperture was set to F6.3 (within the lens’s optimal mid-range) for stable image quality and consistent exposure. Focal length was adjusted to get the best composition for each scene. Camera triggering was performed via Sony’s Imaging Edge mobile app over Wi-Fi, avoiding unwanted shadows and sensor misalignment caused by manual triggering.

\subsubsection{Programmable motion control} Scene motion was precisely controlled using a Kessler CineDrive shuttle dolly system, enabling horizontal translation and three-axis rotation (see Fig.~\ref{fig:viewgeo} (Left)). Start and end points were manually adjusted to ensure that all objects stayed visible and within frame throughout the recorded sequence. This programmable motion setup supports linear, rotational, and angled trajectories -- representative of motion found in surveillance, robotics, and cinematography -- making the dataset practical and representative of real dynamic environments.

% ------------------------------------------------
%%\vspace{-0.1cm}
\subsection{Ground truth generation}
\label{ssec:gtgen}
%%\vspace{-0.1cm}
%The Imaging Edge and Kessler kOS application software operate independently on different Wi-Fi networks; this results in synchronisation issues and can lead to  frame mis-alignment between any two identical sequences.  In addition to camera triggering, further 
The primary sources of misalignment are the mechanical tolerances of the shuttle dolly and the acceleration of the control motor. Even with careful setup, frames from repeated identical motion profiles may be slightly misaligned, with misalignment increasing at higher speeds. In our CineDrive tests, the maximum observed misalignment was 20 pixels, only about 1\% of the frame width (1,920$\times$1,080 pixels).

We preserved the original low-light videos and avoided warping or non-rigid registration to prevent altering noise characteristics, motion blur, or other low-level features~\cite{8451755}. Each low-light frame was paired with a pseudo ground truth frame via our alignment procedure, which relies on repetition-and-selection rather than computational correction. \textbf{\textit{This approach relies on the principle that a very close pseudo reference is sufficient for supervised training}}, with 99.24\% of the selected pairs achieving $<1$ pixel alignment error, supporting effective supervised training (Section~\ref{sec:experiments}).

%as confirmed by our experimental results (Section~\ref{sec:experiments}), where 99.24\% of data alignment is functionally equivalent to perfect ground truth for network training. We avoided warping or non-rigid registration, which can alter low-level features or introduce artifacts \cite{8451755}.

\subsubsection{Repetition strategy} The ground truth videos were recorded under normal lighting. To minimize misalignment with the low-light videos, we repeated the dolly movement, creating three normal-light videos for slow motion and five for fast motion. These repetitions introduced small pixel shifts, ensuring that at least one recording was closely aligned with each low-light video. In most cases, the same normal-light recording provided the best match throughout the sequence; however, in some cases (e.g., when the dolly changed direction), the optimal match switched between repetitions.

\subsubsection{Alignment procedure} 
\begin{figure}[t!]
  \centering
    \includegraphics[trim={15cm 1cm 0  0}, clip, width=\columnwidth]{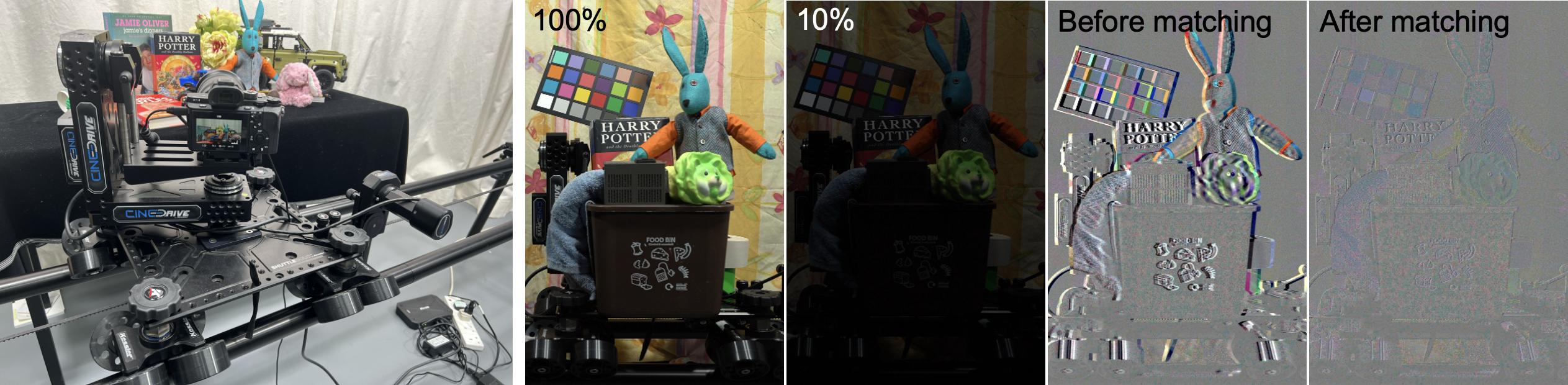} %\vspace{-4mm}
   \caption{Moving bunny scene with static background showing pixel value difference between the normal-light frame and the adjusted low-light frame before and after alignment (gray = zero error).}%\vspace{-4mm}
  \label{fig:bunny2}
\end{figure}

To perform precise frame alignment, we first cropped a 100-pixel border from each frame to remove peripheral background and camera boundary effects. For each low-light frame $J_t$, we applied histogram matching to enhance comparability with a normal-light candidate frame $I^m_n$ from video $m$ at frame $n$, expressed as $\hat{J}_t =  \mathcal{H}(J_t, I^m_n)$, where $\mathcal{H}$ is a histogram-matching function~\cite{SHEN20071161}. We used 256 bins for 8-bit images to ensure full coverage of the intensity range. %Histogram matching was used solely for alignment scoring, not for model training, to ensure the training data remained unaltered. 
The best-aligned reference frame was determined by minimizing the mean absolute error (MAE) between $\hat{J}_t$ and $I^m_n$, within a temporal search window of $n \in [t-N_w, t+N_w]$, where $N_w=3$. Across all repetitions $m \in [1, N_v]$, where $N_v$ is the number of repeated captures, the pair $(m,n)$ with the lowest MAE was selected. Fig.~\ref{fig:bunny2} shows the error maps between normal-light and low-light frames before and after alignment. 
This procedure performs only repetition and selection. Histogram matching is employed only to score candidate matches under different illumination levels and is never applied to the training data. No pixel- or region-level modification, warping, or non-rigid registration is applied to either the low-light inputs or the selected normal-light references; all video pairs remain original camera recordings.

\subsubsection{Alignment accuracy} 
\begin{figure}[t]
    \centering
    \includegraphics[width=\linewidth]{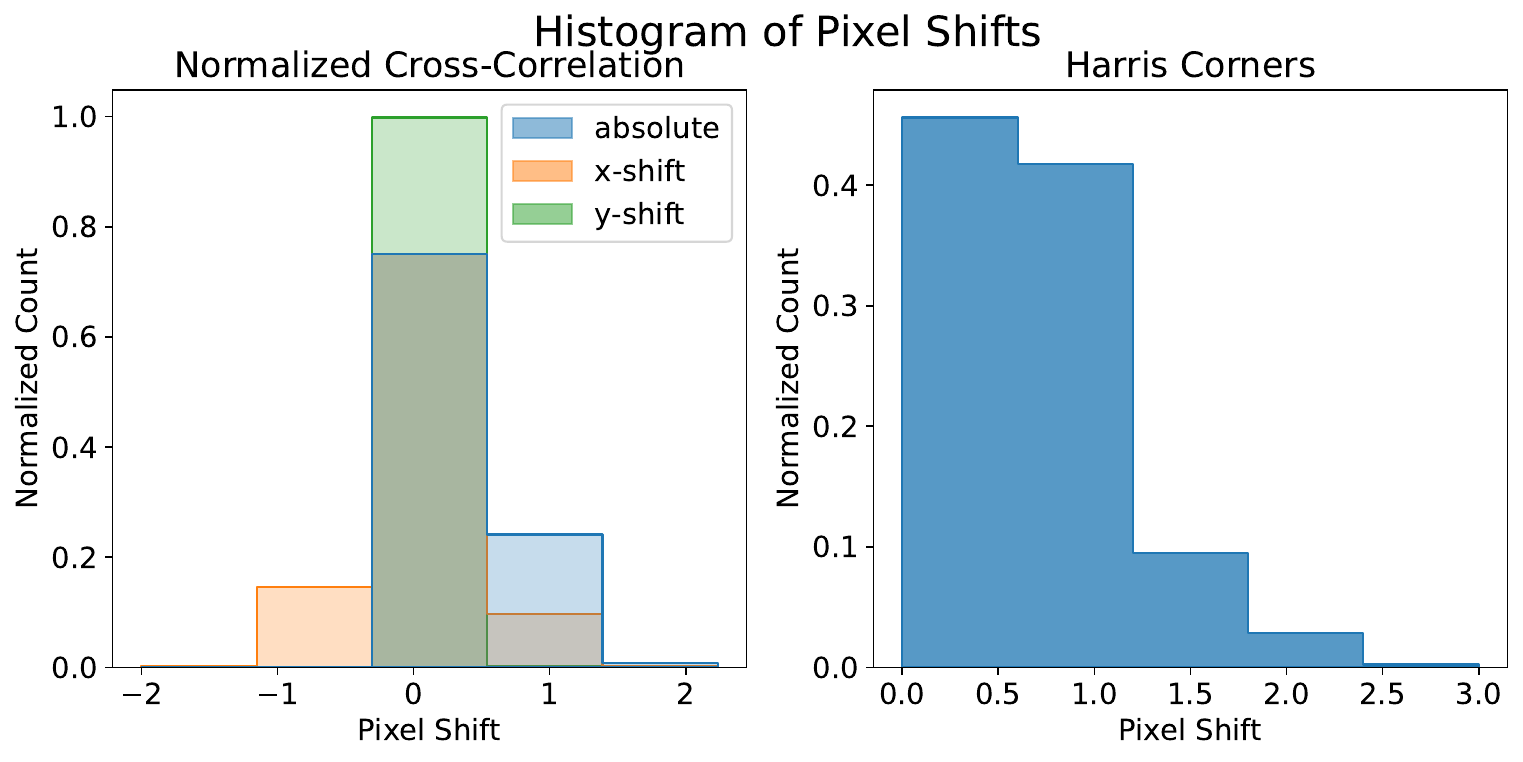}%\vspace{-4mm}
    \caption{Distributions of the pixel shifts for verifying our alignment technique. Left and right sub-figures show the distributions for the cross-correlation method and the corner detection method, respectively. The cross-correlation sub-figure also shows the pixel shift distributions for the $x$ and $y$ directions.}
   \label{fig:alignment_histograms}%\vspace{-4mm}
\end{figure}

We verified the accuracy of our alignment procedure using two complementary methods, with pixel-shift distributions shown in Fig.~\ref{fig:alignment_histograms}. First, we used Normalized Cross-Correlation (NCC)~\cite{sarvaiya2009ncc}, which compensates for illumination differences and is robust for template matching. Using an FFT-based implementation for efficiency, we obtained an average alignment error of 0.256~pixels with a standard deviation of 0.451~pixels, with 99.24\% 
of alignments achieving $<1$ pixel absolute alignment error. Second, we used Harris Corner Detection~\cite{harris1988cornerdet} on frames containing a visible color card. To improve corner detection in low-light images, we applied histogram matching and Otsu’s thresholding~\cite{otsu1979threshold} to isolate the white square, then matched corners across frames using the Hungarian algorithm~\cite{kuhn1955hungarianalg} with Euclidean distance as the cost metric. This method yielded an average displacement of 0.619~pixels with a standard deviation of 0.610~pixels, confirming the alignment accuracy observed with NCC. Supplementary Section~S-I analyzes the remaining 0.76\% non-sub-pixel cases, which mainly arise from abrupt motion changes, reflective surfaces, or texture-poor regions.

%------------------------------------------------------
  \begin{figure*}[t!]
   % %\vspace{-1.3cm}
    \centering
    \includegraphics[width=\linewidth]{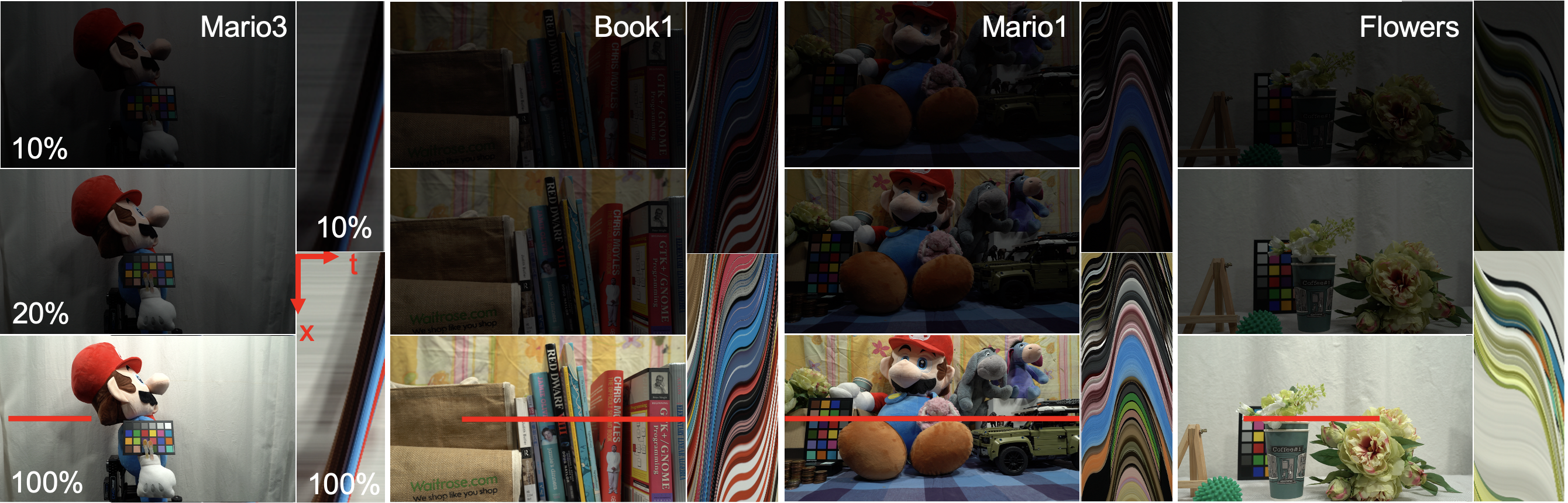} %\vspace{-4mm}
  \caption{Scene examples with varying light levels and different motion profiles as shown in x-t planes. The length of the red lines across the normal-light frames represents the height of the x-t planes.}
  \label{fig:motions}
 %\vspace{-4mm}
 \end{figure*}

\subsection{BVI-RLV dataset}
\label{ssec:dataset}
This work focuses mainly on achieving full frame-wise registration, which is essential for generating reliable paired data in low-light video enhancement. Accurate alignment between distorted and ground truth videos is particularly challenging in outdoor environments when relying on natural illumination instead of ND filters. To ensure precise alignment, we therefore captured videos under controlled indoor lighting while introducing extensive variations in object configuration, camera motion, and depth of field to maintain meaningful content diversity. Objects were recorded from different angles, positions, focal lengths, and motion trajectories -- both in-focus and out-of-focus (partially similar to varying depth of field) -- to decouple object appearance from fixed scene layouts and reduce the risk of overfitting to specific spatial configurations, ISO settings, illumination levels, or camera types. %This approach enables more generalizable model training and facilitates rigorous analysis of factors influencing performance. 
This controlled setup ensures precise registration while preserving content, texture, motion, and illumination diversity,  enabling rigorous analysis of performance factors (see Section~\ref{sec:experiments}). By prioritizing motion diversity and accurately registered video pairs, BVI-RLV avoids the misalignment and optical artifacts common in less-curated outdoor capture, yet generalizes effectively to outdoor scenes (see Section~\ref{ssec:outdoor}).
% The benefits of this approach are evident from the performance of models trained on our dataset when evaluated on multiple datasets (see Section~\ref{sec:experiments}). Model performance in low-light conditions is primarily driven by motion diversity and accurately registered video pairs, both of which are integral features of the proposed dataset. Notably, our models also perform well on outdoor scenes (see Section~\ref{ssec:outdoor}).

In total, the dataset contains 31,800 paired frames with full HD resolution, where the 100\% light level serves as the normal-light ground truth. We captured both static backgrounds with moving objects and fully dynamic scenes. For static backgrounds, objects were placed on a dolly while the camera remained on a tripod. For dynamic content, the camera was mounted on a dolly, creating both linear and nonlinear motions. Three camera positions and movements relative to the dolly’s direction were configured: i) \textit{parallel} (camera axis perpendicular to the dolly movement), ii) \textit{angled} (camera axis at a 30-degree angle to the dolly movement), and iii) \textit{rotational} (camera rotation in the yaw axis while the dolly moves). Dolly speeds were either constant or varied after clipping, denoted as \textit{linear} and \textit{nonlinear}, respectively. The total number of frames per scene varied with motion speed. Each scene included a Datacolor SpyderCheckr24 card for camera calibration and as a white balance target, providing homogeneous reference patches. Full dataset documentation is provided in the accompanying datasheet on our project website.

\begin{figure}
    \centering
    \includegraphics[trim={0 0 19.8cm  0}, clip, width=\columnwidth]{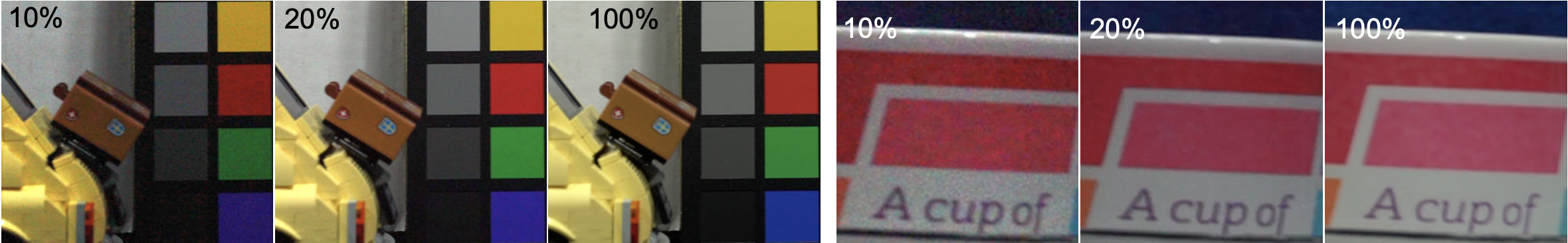}
    \includegraphics[trim={22.15cm 0 0  0}, clip, width=\columnwidth]{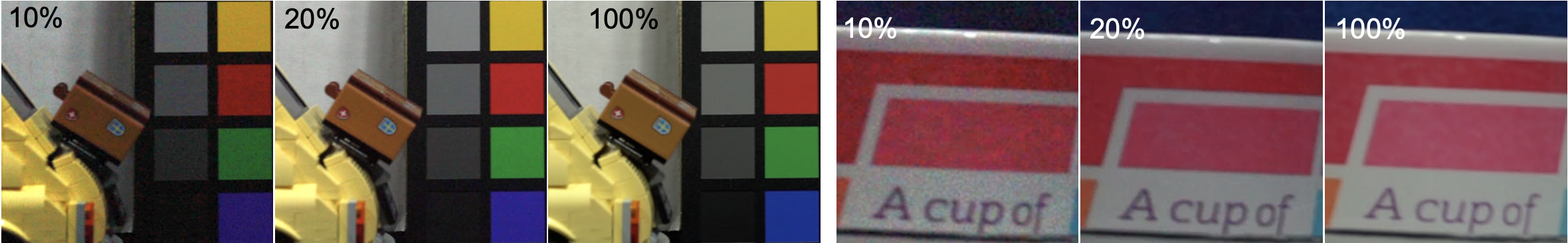} %\vspace{-4mm}
    \caption{Cropped images of Lego (top row) and Kitchen (bottom row) scenes at 350$\times$350 and 140$\times$140 pixels, respectively, with histogram matching to the reference (normal light) to visualize noise at different levels of light.}% (From left to right) Light levels of 10\%, 20\%, and normal light (100\%).}
    \label{fig:crop_histadjust}
    %\vspace{-5mm}
\end{figure}

Fig.~\ref{fig:motions} shows representative motion patterns through x-t planes. Mario3 contains linear foreground motion with a static background, Book1 transitions from fast to slow and back to fast, while Mario1 and Flowers show nonlinear motion. Fig.~\ref{fig:crop_histadjust} visualizes low-light noise after histogram matching to the 100\% reference. Extensive color correction is unnecessary, as the standard color palette included in each scene aids the training process. The magnified Kitchen crop shows motion blur, where the letters (`A cup of') appear thicker under low light. Reflective surfaces are included in Lego, Kitchen, Wires, Books1, and Books4.

% Fig.~\ref{fig:motions} shows various captured motions. The x-t planes of the four example videos reveal different movement patterns. Mario3 shows linear motion of the foreground object with a static background. The dynamic scene Book1 transitions from fast motion to slow and back to fast. Mario1 exhibits nonlinear movement from left to right and back. The Flowers scene also features nonlinear movement. Fig.~\ref{fig:crop_histadjust} illustrates the noise present in the low-light scenes (for visualization, histogram matching was applied to 100\% lighting). Extensive color correction is unnecessary, as the standard color palette included in each scene aids the training process. The right image group shows a magnified, cropped region of the mug in the Kitchen scene. The letters (`A cup of') in the low-light video appear thicker than in the normal-light one due to motion blur. In addition, reflective surfaces are included in the following scenes: Lego, Kitchen, Wires, Books1, and Books4.

\begin{table}[t!]
    \centering
    \caption{Standard deviation of high-frequency magnitudes across different wavelet decomposition levels. The best results are highlighted in \textbf{bold}, and the second best are \underline{underlined}}
    %\footnotesize
    %\setlength{\tabcolsep}{10pt}
    %\renewcommand{\arraystretch}{1.1}
    %\resizebox{\columnwidth}{!}{%
    \begin{tabular}{@{}c|ccccc|c@{}}
    \toprule
   \multirow{2}{*}{Dataset} & \multicolumn{5}{c|}{Wavelet level} & \multirow{2}{*}{Avg.} \\
         & 1 & 2 & 3 & 4 & 5 \\ \hline
        DRV & \underline{0.011} &    \textbf{0.037}  &   \textbf{0.091}  &   \underline{0.222}  &   0.517 & 0.176\\
SDSD & 0.006   &   0.025   &   0.080  &    0.216    &  \underline{0.557} & \underline{0.177}\\
 DID & \textbf{0.013} &   0.033 &  0.083 &  0.199 &  0.466 & 0.159
 \\
BVI-RLV (ours) & 0.010 &   \underline{0.034} &  \underline{0.089} &  \textbf{0.230} &  \textbf{0.566} & \textbf{0.186} \\
    \bottomrule
    \end{tabular}
  %  }
    \label{tab:contentmeasure}
    %\vspace{-4mm}
\end{table}

% \subsubsection{Content variety comparison} 
Moreover, we investigated the variety of content within the datasets, including homogeneous areas, textures, and structures. To assess this, we applied the discrete wavelet transform to extract high frequencies from the sequences. Specifically, we transformed the frame in the middle of each video into five wavelet decomposition levels. Level 1 provides finer details, while higher levels expose more structures. We measured the variety of content using the standard deviations of the high-frequency magnitudes at each decomposition level, as shown in Table \ref{tab:contentmeasure}. Our dataset achieves the highest average standard deviation across all levels, indicating the most diverse content overall. Together with DRV, it generally covers a broader range of content than DID and SDSD. Our dataset exhibits stronger structural information, which benefits deblurring tasks and produces perceptually sharper results.

% ------------------------------------------------
\begin{figure*}[t!]
    \centering
    \includegraphics[width=1.\linewidth]{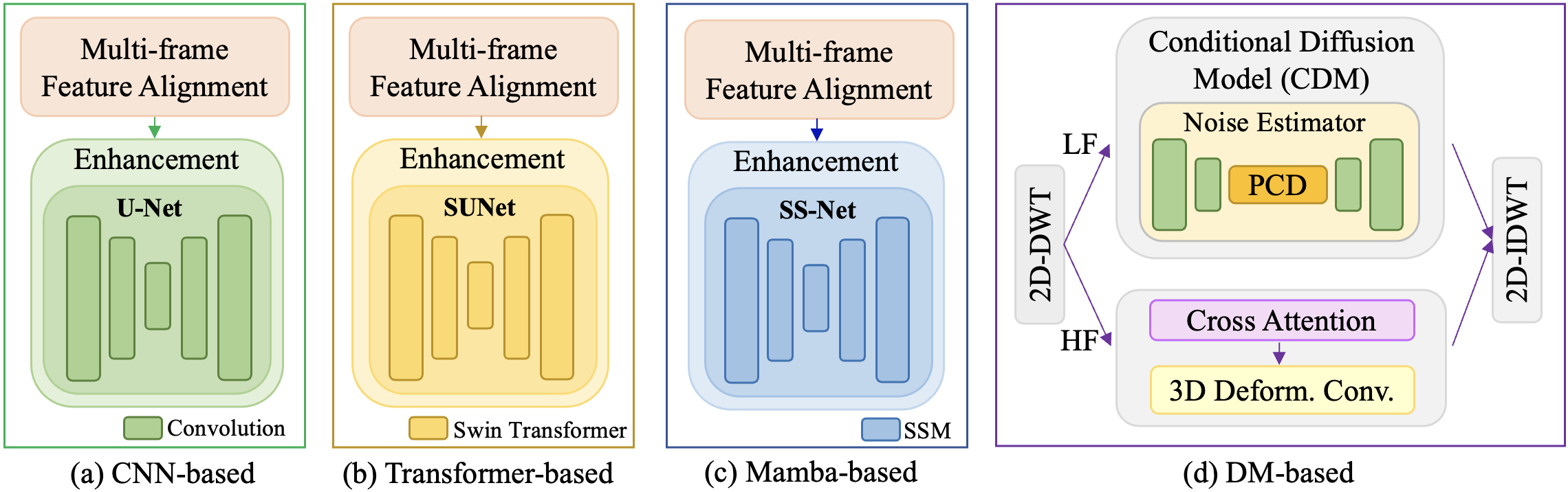} %\vspace{-6mm}
    \caption{Main architectural components of the four different baselines, (a) CNN-based, (b) Transformer-based, (c) Mamba-based, and (d) DM-based.}
    \label{fig:arch}
    %\vspace{-4mm}
\end{figure*}

\section{Baseline methods for benchmarking}
\label{sec:Benchmarks}

To enable fair and reproducible evaluation, we provide four standardized baselines based on widely used architectures. These are not intended as methodological contributions, but as consistent implementations for transparent benchmarking across datasets. As evidenced in Section~\ref{sec:experiments}, they help confirm that BVI-RLV's superiority is due to data quality rather than method complexity. All experiments use a single NVIDIA RTX 3090 GPU, and all baselines take five input frames. The main architectural components are shown in Fig.~\ref{fig:arch}.

% To facilitate fair comparison and reliable evaluation on our dataset, we provide four baseline methods built on widely used architectures. These baselines are not intended as methodological contributions but rather as standardized implementations to ensure reproducible and transparent evaluation across different datasets, especially given the limited availability of video-based enhancement methods. Although these architectures are relatively straightforward, they establish a consistent reference point for benchmarking our proposed dataset and confirming that its superiority is due to data quality rather than method complexity, as evidenced in Section~\ref{sec:experiments}. All experiments are conducted on a single NVIDIA GeForce RTX 3090 GPU to ensure reproducibility. The main architectural components are illustrated in Fig.~\ref{fig:arch}. To ensure comparability, we set the input to five frames for all baseline methods. 

The first three baselines follow a unified two-stage framework combining multi-frame feature alignment with U-Net architectures, differing only in their core modules. This design allows us to benchmark performance across representative paradigms while maintaining comparability. Specifically, (a) \textbf{CNN-based} employs Pyramid, Cascading, and Deformable (PCD) alignment~\cite{Wang:EDVR:2019} with standard U-Net, forming a lightweight version of EDVR; (b) \textbf{Transformer-based} replaces the CNN backbone with Swin Transformer~\cite{liu2021swin} for hierarchical local-global feature capture; (c) \textbf{Mamba-based} substitutes Swin Transformer with state space models~\cite{liu2024vmamba}, enabling linear-complexity long-sequence modeling; and (d) \textbf{DM-based}, inspired by~\cite{Jiang:Low:2023}, adapts a conditional diffusion model (CDM)~\cite{song2020denoising} to operate low-frequency (LF) information in the wavelet domain to reduce memory and complexity, and process high-frequency (HF) information separately using cross attention and exploit spatiotemporal information with 3D deformable convolutions.

% ------------------------------------------------
% \setlength{\tabcolsep}{4pt}
\begin{table*}[!ht]
    \centering
      \caption{Quantitative cross-dataset performance comparison (PSNR$\uparrow$/SSIM$\uparrow$/LPIPS$\downarrow$) of the models trained with and tested on different datasets. The reported results are the unweighted average of the test sets from the \texttt{four} datasets. \textbf{Bold} and  \underline{underlined} indicate the best and second-best performance of each method. Image-based models are marked in \textcolor{gray}{gray} %mo-CGAN is a modification of CGAN \cite{Wang:hight:2018} to accept 5-frame input.
    }
   % \small
   % \setlength{\tabcolsep}{16pt}
   % \renewcommand{\arraystretch}{1.1}
    \begin{tabular}{@{}l|cccc@{}}
    \toprule
        \centering
        % \multirow{2}{*}{}
        % Trained with 
        Dataset & DRV \cite{Chen_2019_ICCV} & SDSD \cite{wang2021sdsd} & DID \cite{Fu:dancing:2023} & {BVI-RLV (ours)}  \\ 
        \hline
      %  Method & \multicolumn{4}{c}{Without postprocessing} \\ \hline
        %UNet$^*$  \cite{ronneberger2015unet} & \textbf{17.52}/\textbf{0.537} & 13.31/0.457 & 16.79/0.490 & 17.38/0.524\\
        % Metrics
        % \cmidrule(r){2-5} 
        Method & PSNR/SSIM/LPIPS & PSNR/SSIM/LPIPS & PSNR/SSIM/LPIPS & PSNR/SSIM/LPIPS  \\ \hline
        \textcolor{gray}{RIDNet} \cite{Anwar:Real:2019} & \textcolor{gray}{18.43/0.678/0.340} & \textcolor{gray}{18.15/0.675/0.345} & \textcolor{gray}{\textbf{20.88}/\textbf{0.762}/\textbf{0.237}} &\textcolor{gray}{ \underline{19.69}/\underline{0.756}/\underline{0.261}}\\ 
        \textcolor{gray}{SwinIR} \cite{Liang:SwinIR:2021} & \textcolor{gray}{\underline{15.22}/\underline{0.707}/0.421} & \textcolor{gray}{11.00/0.436/0.533} & \textcolor{gray}{11.58/0.588/\underline{0.410}} & \textcolor{gray}{\textbf{17.54}/\textbf{0.733}/\textbf{0.348}}\\
        mo-CGAN \cite{Wang:hight:2018}  & 15.84/0.456/0.380 & 14.25/0.431/0.481 & \underline{15.87}/\underline{0.512}/\underline{0.377} & \textbf{17.41}/\textbf{0.557}/\textbf{0.302}  \\
        EDVR \cite{Wang:EDVR:2019} & 15.73/0.598/0.354 & 16.20/0.640/0.330 & \underline{19.53}/\underline{0.701}/\underline{0.275} & \textbf{20.12}/\textbf{0.756}/\textbf{0.214} \\
        SMOID-Net \cite{Jiang:learn:2019} & \underline{17.66}/0.541/0.379 & 14.92/0.501/0.401 & 17.62/\underline{0.554}/\underline{0.323} & \textbf{17.98}/\textbf{0.621}/\textbf{0.280} \\
        SDSD-Net \cite{wang2021sdsd} &  16.31/0.587/0.366 &  16.08/0.569/0.359 & \textbf{19.05}/\underline{0.652}/\underline{0.291} & \underline{18.05}/\textbf{0.690}/\textbf{0.257} \\ \hline
        CNN-based model & 15.00/0.492/0.422 & 17.07/0.677/0.356 & \underline{18.91}/\underline{0.725}/\textbf{0.305} & \textbf{19.50}/\textbf{0.757}/\underline{0.316}\\
        Transformer-based model & 14.19/0.424/0.424 & 16.93/0.608/0.475 & \underline{19.61}/\underline{0.714}/\underline{0.385} & \textbf{20.64}/\textbf{0.765}/\textbf{0.243} \\
        Mamba-based model& \textbf{22.95}/\underline{0.784}/\underline{0.195} & 19.91/0.751/0.281 & 18.74/0.738/0.238 & \underline{22.93}/\textbf{0.820}/\textbf{0.146} \\
        DM-based model& 19.56/0.575/0.350 & \underline{21.00}/\underline{0.746}/0.292 & 19.44/0.683/\underline{0.199} & \textbf{22.20}/\textbf{0.773}/\textbf{0.175} \\
        %BVI-Mamba  & - & - & -& \textbf{22.93}/\textbf{0.821}/\textbf{0.147} \\
        
        %\hline
        %Method & \multicolumn{4}{c}{Postprocessing with histogram matching to a reference} \\
        %\hline
        %SwinIR$^*$ \cite{Liang:SwinIR:2021} & 29.70/0.853 & 27.60/0.800 & 30.53/0.883 & \textbf{31.97}/\textbf{0.915}\\
        %mo-CGAN \cite{Wang:hight:2018}  & 31.04/0.853 & 29.35/0.820 & 33.68/0.868 & \textbf{34.66}/\textbf{0.906}  \\
    \bottomrule
    \end{tabular} 
  
    \label{tab:alltest}
    %\vspace{-0.2cm}
\end{table*}

\section{Experiments and discussion}
\label{sec:experiments}

% This section examines the importance of fully registered video pairs in the development of low-light video enhancement methods. We compared our proposed dataset with existing ones by using them to train several state-of-the-art methods for low-light enhancement and video restoration, as well as our benchmarks.
% In all experiments using our dataset, we randomly selected 32 scenes for training and allocated the remaining scenes for testing. These were fixed for all experiments.

\subsection{Comparisons with other datasets}
% -----------------------------------
We evaluated the training effectiveness of our datasets using various low-light video enhancement methods, including SMOID-Net~\cite{Jiang:learn:2019}, SDSD-Net~\cite{wang2021sdsd}, our CNN-, Transformer-, Mamba-, and DM-based models, and the video restoration method EDVR~\cite{Wang:EDVR:2019}. We also extended CGAN~\cite{Wang:hight:2018} to support multiple input frames, referring to it as mo-CGAN. For applications requiring faster, lower-computation processing, we further assessed image-based methods such as RIDNet~\cite{Anwar:Real:2019} and SwinIR~\cite{Liang:SwinIR:2021}.

We compared models trained on our datasets with those trained on existing ones: DRV~\cite{Chen_2019_ICCV} (publicly available only in 8-bit RGB format), SDSD~\cite{wang2021sdsd}, and DID~\cite{Fu:dancing:2023}. Dataset splits followed the original papers: 73:27 for DRV, 84:16 for SDSD, and 326:87 for DID. Our dataset was divided 80\%-20\% (32 scenes for training). The training parameters were also aligned with those outlined in the original papers, ensuring consistency across all datasets for a fair comparison. Unless specified otherwise, we trained each model with 200k iterations.

\begin{figure*}[!th]
\centering
    \includegraphics[width=\textwidth]{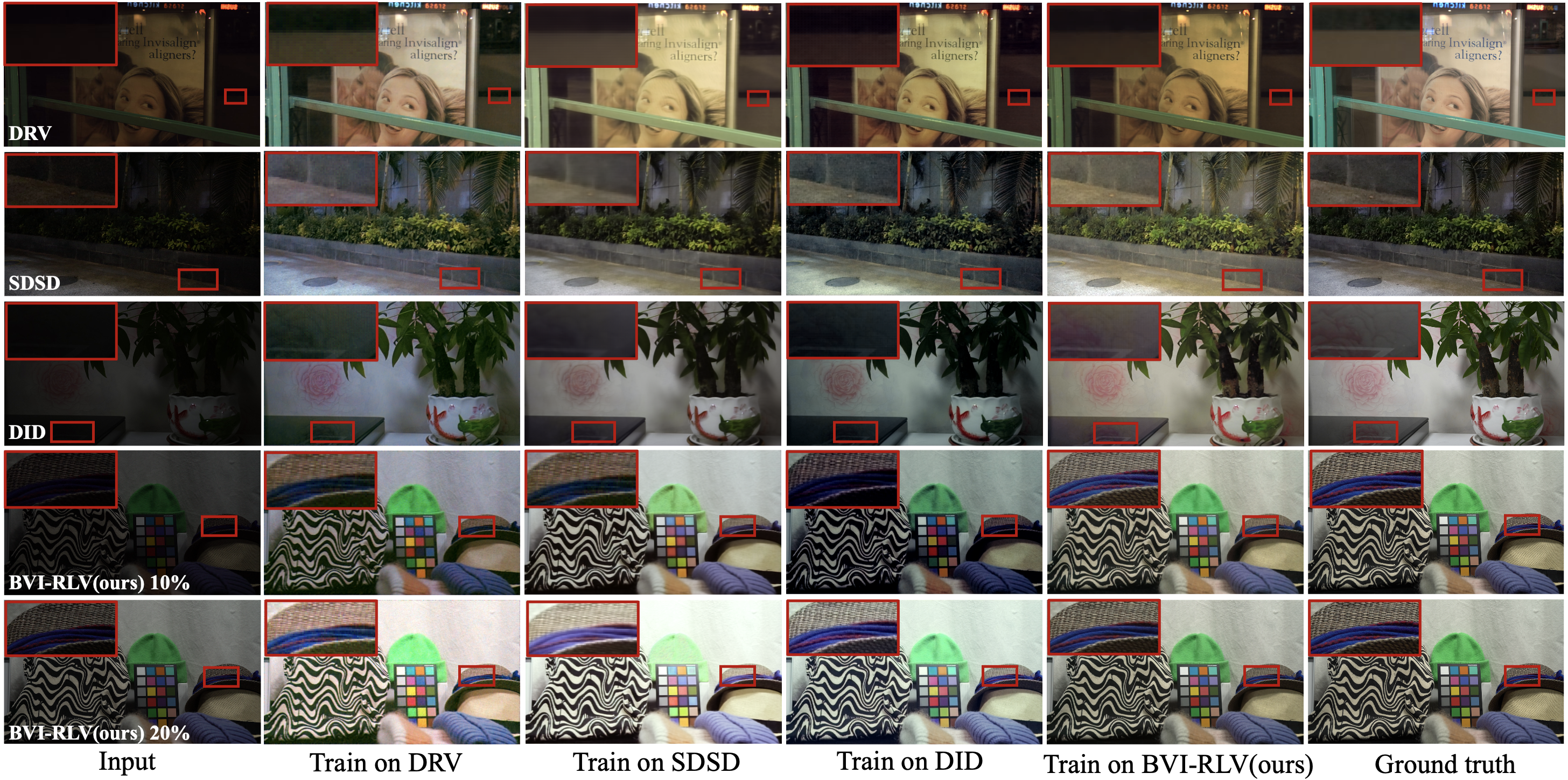} %\vspace{-7mm}
    \caption{Qualitative results of the DM-based model trained on different datasets, and tested on different datasets. The test results for the DRV, SDSD, DID, and our dataset (10\% and 20\% light levels) are displayed from top to down.}\label{fig:visualresults} %\vspace{-3mm}
\end{figure*}

Table~\ref{tab:alltest} reports the unweighted average cross-dataset generalization performance across four datasets. Models trained on BVI-RLV (ours) consistently outperform those trained on DRV, SDSD, and DID, in agreement with the visual comparisons in Fig.~\ref{fig:visualresults}. Across all test sets, the BVI-RLV-trained model produces cleaner, sharper results with improved noise suppression and detail restoration. On the SDSD test set, it yields sharper textures and edges even compared with models trained on SDSD. On DID, it correctly reconstructs the wallpaper reflection on the laptop, whereas other models misinterpret it as a shadow. On the BVI-RLV test sets, it better preserves fabric and color fidelity while avoiding artifacts observed in other training settings. We attribute these gains to full registration and the dataset’s diverse coverage of objects, textures, and motion. 
The grouped statistics in Supplementary Section~S-II further confirm this trend: the BVI-RLV-trained Mamba model achieves the best PSNR/SSIM/LPIPS across motion speed, movement type, motion complexity, and illumination-level subsets, showing that the gains are not driven by favorable cases.

Finally, each dataset was captured under different illumination and white-balance conditions, so cross-dataset color shifts are expected. BVI-RLV was recorded at 6500~K (daylight), hence models trained on it may produce slightly warmer/darker outputs when evaluated on datasets with higher color temperatures or lower light levels. This color variation does not affect the observed improvements in sharpness, noise suppression, or detail fidelity.

\begin{figure*}[t!]
    \centering
    \includegraphics[width=1.\linewidth]{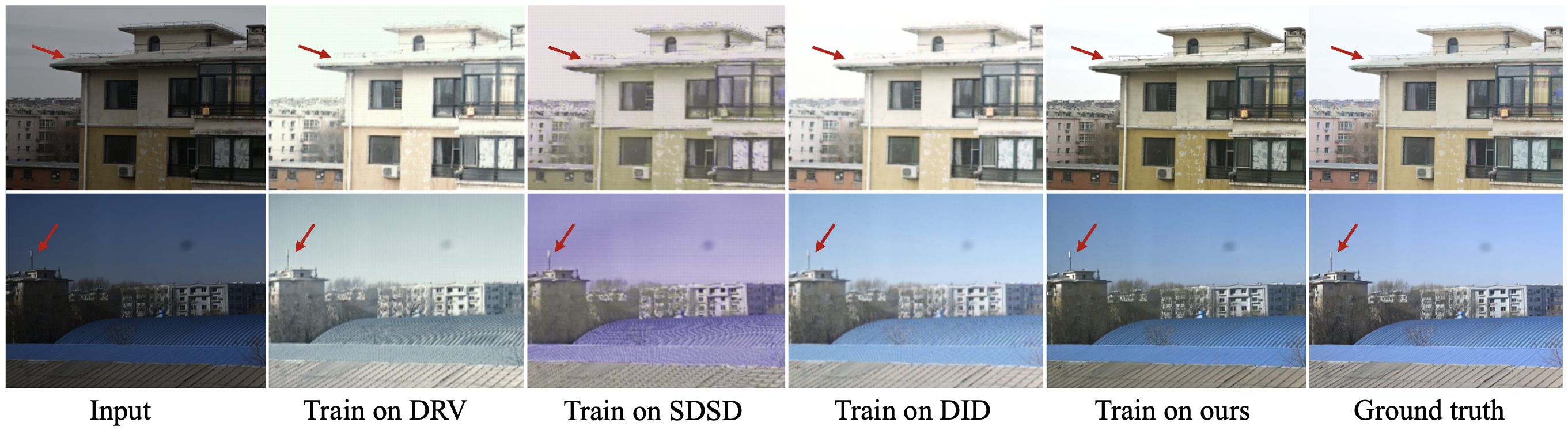} %\vspace{-6mm}
    \caption{Qualitative results of the transformer-based model trained on different datasets, and tested on the outdoor scenes of DID.}
    \label{fig:outdoor} %\vspace{-4mm}
\end{figure*}

%----------------------------------------------------------------------
\begin{table}[t!]
\centering
\caption{Quantitative performances of DM-based model trained with different datasets and tested on SDSD outdoor sets. \textbf{Bold} and  \underline{underlined} indicate the best and second-best performance} %\vspace{-3mm}
\begin{tabular}{@{}c|ccc@{}}
    \toprule  
        Trained with & PSNR&SSIM&LPIPS \\  \hline
        %SDSD (no domain shift)  & \textbf{25.37}&\textbf{0.808}&0.288  \\
        DRV dataset  & 10.47&0.386&0.338  \\ 
        DID dataset  & \underline{20.37}&\underline{0.609}&\underline{0.237}  \\ 
        BVI-RLV dataset (ours) & \textbf{21.96}&\textbf{0.708}&\textbf{0.165}  \\
    \bottomrule
    \end{tabular}

\label{tab:sdsdout}%\vspace{-6mm}
\end{table}
\subsection{Evaluation results on outdoor scenes.} 
\label{ssec:outdoor}

\begin{figure}[t!]
    \centering
    \includegraphics[width=\columnwidth]{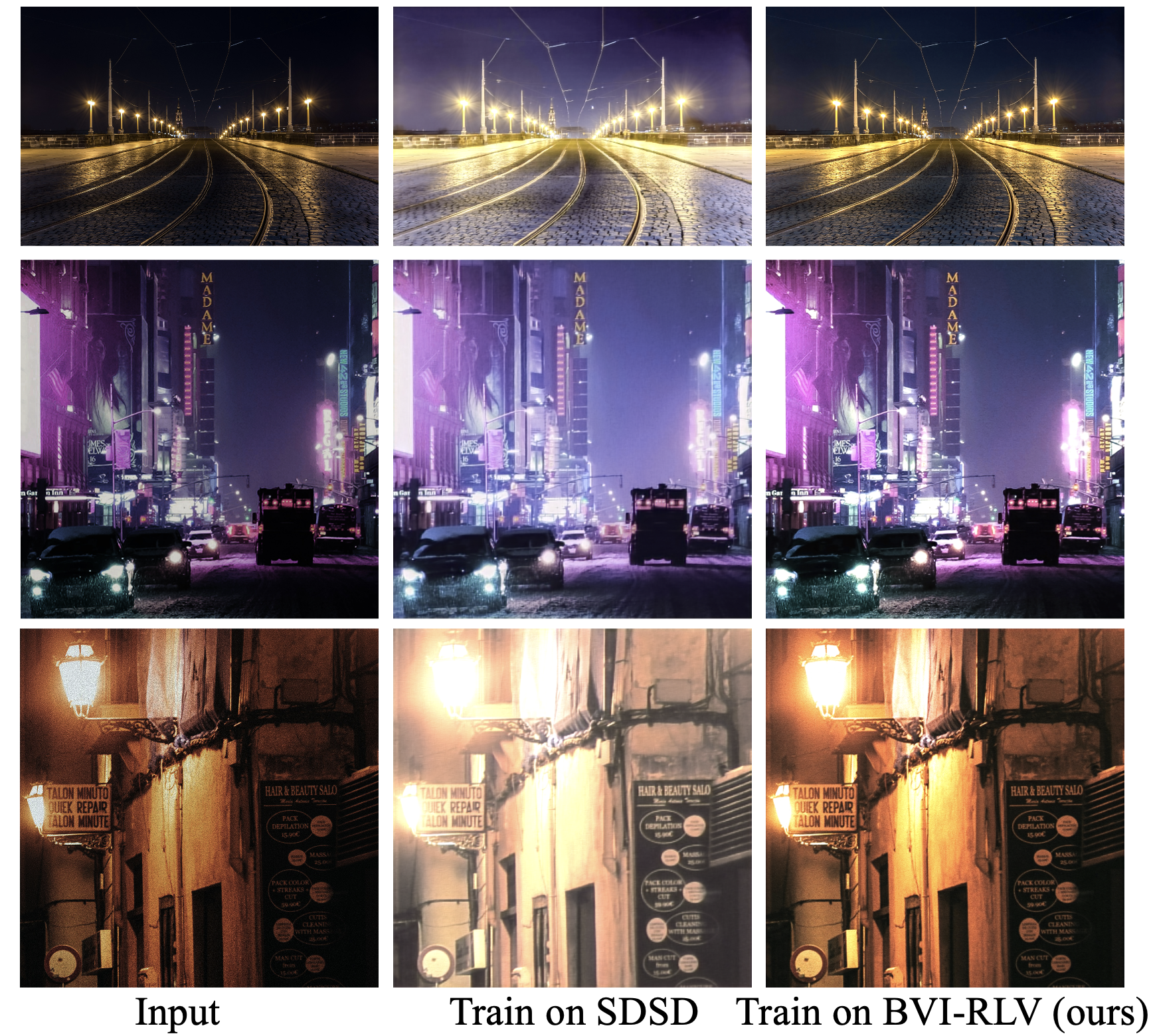} %\vspace{-6mm}
    \caption{Qualitative results of the DM-based model trained on different datasets, and tested on public outdoor scenes with local light sources.}
    \label{fig:public_outdoor} %\vspace{-4mm}
\end{figure}

Models trained on BVI-RLV also generalize effectively to outdoor scenes, as shown in Fig.~\ref{fig:visualresults} (SDSD test video) and Fig.~\ref{fig:outdoor} (DID test video). We also provide an objective evaluation on the SDSD outdoor video dataset using the DM-based model trained on different datasets, with results summarized in Table~\ref{tab:sdsdout}. The model trained on BVI-RLV achieves the best overall performance, both qualitatively and quantitatively, producing outputs with stronger structure and comparable fine details (zoom in on Fig.~\ref{fig:outdoor} for better visualization). Finally, we tested models on real night scenes\footnote{Row 1 uses Pixabay image by StockSnap; rows 2 and 3 use Unsplash images by Valentin and Eddie \& Stigson, respectively.} with non-uniform local light sources. As shown in Fig.~\ref{fig:public_outdoor}, the BVI-RLV-trained model preserves realistic texture and detail in dark regions while avoiding overexposure, unlike the SDSD-trained model. Additional qualitative outdoor results are provided in Supplementary Section~S-III.

 \subsection{Downstream task performance} 
We further evaluate the practical value of BVI-RLV on ELVIS-S~\cite{lin2026elvisenhancelowlightvideo}, an independent manually annotated real-world low-light VIS dataset. The same Mamba-based enhancement model is trained on different datasets, and the enhanced frames are evaluated using a frozen MinVIS~\cite{huang2022minvis} model. Frames enhanced by the BVI-RLV-trained model achieve the best overall results for both video instance segmentation and per-frame object detection, improving mask AP by 5.28 points and box AP by 3.40 points over the second-best training dataset. This indicates better preservation of instance-level cues after enhancement. Detailed implementation, quantitative results, and visual examples are provided in Supplementary Section~S-IV.

% ------------------------------------------------------------
\subsection{Ablation study}
% \subsection{Dataset evaluation}

\begin{table}[t]
\centering
\caption{Quantitative comparison of results with/without using temporal (T) and motion (M) information. The best results are highlighted in \textbf{bold}, and the second best are \underline{underlined} }
%\small
%\setlength{\tabcolsep}{26.5pt}
%\renewcommand{\arraystretch}{1.1}
\begin{tabular}{@{}cccc@{}}
\toprule
        Use &  Methods &  PSNR &  SSIM  \\ 
        \hline
        - &  Single-Input/Single-Output  & 21.78 & 0.770\\
        Temporal cues &  Multi-Input/Single-Output & \underline{24.48} & \underline{0.844}\\
        Temporal+Motion cues &  Multi-Input/Multi-Output & \textbf{25.23} & \textbf{0.877}\\
    \bottomrule
\end{tabular}
%\vspace{-5mm}
\label{tab:studymotion}
\end{table}

\subsubsection{Importance of dynamic video data} \label{ssec:dynamicvideo} We study the role of temporal dynamics in low-light video enhancement, focusing on why motion-rich video datasets outperform image-based or static video datasets such as Chen et al.~\cite{Chen_2019_ICCV}. We train U-Net models under three settings: single-frame input/output (no temporal or motion cues), multi-frame input/single-frame output (temporal only), and multi-frame input/output (temporal + motion). We extend U-Net to produce multiple outputs and compute losses against corresponding ground-truth frames, enabling full exploitation of motion during training. 

As shown in Table~\ref{tab:studymotion}, incorporating temporal information significantly improves performance over single-frame inputs, and adding explicit motion cues further boosts PSNR and SSIM. While Chen et al.~\cite{Chen_2019_ICCV} suggest that static video datasets with randomly paired frames can generalize to dynamic scenarios, our results show that such models fail to fully exploit temporal information. These findings indicate that both temporal and motion cues are essential for stable performance in dynamic scenes with rapid illumination or color changes that can otherwise lead to frame-to-frame inconsistency.

% ---------------------------------------------
\subsubsection{Importance of fully registered data}
\label{sssec:fullyregister_exp}
We examine the effect of frame registration on supervised model learning by retraining three representative architectures -- U-Net (CNN)~\cite{ronneberger2015unet}, CGAN (GAN)~\cite{Wang:hight:2018}, and SwinIR (Transformer)~\cite{Liang:SwinIR:2021} -- on our dataset using frames before registration (unregistered) and after registration (registered). Training employed common loss functions with varying sensitivity to spatial misalignment, including $\ell_1$, SSIM, and VGG~\cite{anantrasirichai:AI:2022}. Pixel-wise losses (e.g., $\ell_1$) are highly sensitive to misalignment, while perceptual losses (SSIM, VGG) are more tolerant.

\begin{figure}[t]
  \centering
\includegraphics[trim={0 0 22.56cm  0}, clip, width=\columnwidth]{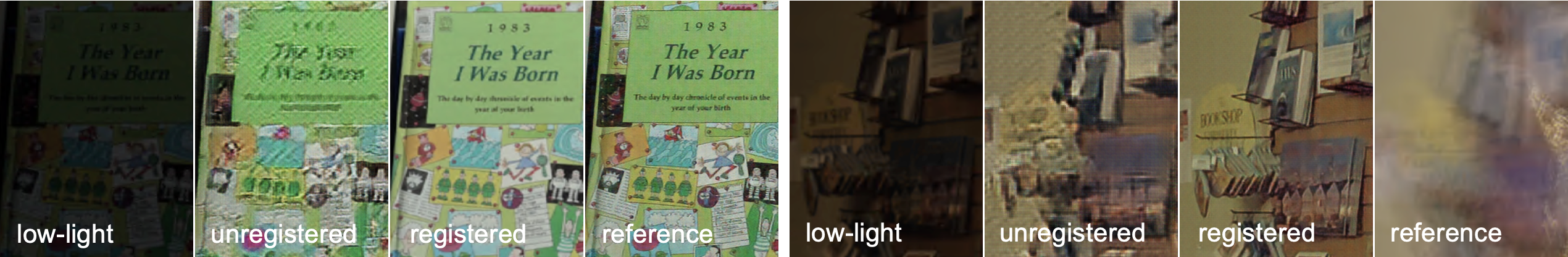}

\includegraphics[trim={22.6cm 0 0  0}, clip, width=\columnwidth]{results/unregistered.png}
   % \caption{Comparison of training CGAN \cite{Wang:hight:2018} with the unregistered and registered datasets. The top and bottom rows show cropped frames from dynamic test scenes from our dataset and DRV dataset, respectively.}
   \caption{Qualitative results of CGAN~\cite{Wang:hight:2018} trained with unregistered and registered datasets. Top: dynamic scenes from our dataset with ground truth. Bottom: DRV scenes using long-exposure frames as brightness references.}
   \label{fig:studyregister}
   %\vspace{-3mm}
\end{figure}

Fig.~\ref{fig:studyregister} presents qualitative results on dynamic scenes from our dataset (top) and DRV (bottom). Since DRV lacks ground-truth dynamic frames, long-exposure images are used as brightness references. Models trained on unregistered data exhibit blur and ghosting due to spatial misalignment, confirming the importance of registration for temporally coherent and perceptually faithful restoration. Table~\ref{tab:studyregister} compares unregistered and registered training. Registration consistently improves performance across architectures and losses, with the largest gains under $\ell_1$ loss (CGAN: +5.85~dB PSNR, +0.160 SSIM). Even perceptual losses benefit, with SwinIR gaining +3.59~dB PSNR under SSIM loss, indicating that even shift-tolerant objectives are improved by precise alignment. Overall, accurate registration is crucial for stable supervised video enhancement.

\begin{table*}[t]
\centering
\caption{Impact of frame registration on model performance across losses. $\Delta$ indicates improvement after registration (Reg.--Unreg.)}
\setlength{\tabcolsep}{4pt}
\renewcommand{\arraystretch}{1.05}
\footnotesize
\resizebox{\textwidth}{!}{
\begin{tabular}{@{}l|ccc|ccc|ccc@{}}
\toprule
\multirow{2}{*}{Use} &
\multicolumn{3}{c|}{$\ell_1$ Loss} &
\multicolumn{3}{c|}{SSIM Loss} &
\multicolumn{3}{c}{VGG Loss} \\
\cmidrule(lr){2-10}
 & Unreg. & Reg. & $\Delta$ &
   Unreg. & Reg. & $\Delta$ &
   Unreg. & Reg. & $\Delta$ \\
\hline
Method & PSNR/SSIM & PSNR/SSIM & PSNR/SSIM & PSNR/SSIM & PSNR/SSIM & PSNR/SSIM  & PSNR/SSIM & PSNR/SSIM & PSNR/SSIM \\
\hline
UNet   & 20.05/0.760 & 20.72/0.761 & +0.67/+0.001 &
         19.45/0.731 & 20.37/0.785 & +0.92/+0.054 &
         19.89/0.752 & 20.94/0.753 & +1.05/+0.001 \\
CGAN   & 20.75/0.702 & 26.60/0.862 & +5.85/+0.160 &
         23.91/0.786 & 26.14/0.861 & +2.23/+0.075 &
         24.60/0.796 & 28.41/0.911 & +3.81/+0.115 \\
SwinIR & 20.69/0.693 & 23.62/0.745 & +2.93/+0.052 &
         18.45/0.677 & 22.04/0.732 & +3.59/+0.055 &
         23.15/0.720 & 25.49/0.794 & +2.34/+0.074 \\
\hline
\textit{Avg. $\Delta$} &
\multicolumn{3}{c|}{\footnotesize +3.21/+0.055} &
\multicolumn{3}{c|}{\footnotesize +1.85/+0.052} &
\multicolumn{3}{c}{\footnotesize +2.07/+0.043} \\
\bottomrule
\end{tabular}}
\label{tab:studyregister}
% %\vspace{-3mm}
\end{table*}

%---------------------------------------------------------

\begin{table}[t!]
    \centering
    \caption{Quantitative results of the CNN-based model trained with different illumination levels. \textbf{Bold} and \underline{underlined} indicate best and second-best performance per test-light setting}
    \small
    \setlength{\tabcolsep}{2pt}
    \renewcommand{\arraystretch}{1.1}
    \begin{tabular}{@{}c|ccc@{}}
    
    \toprule
    Trained with & 10\% & 20\% & Mixed (10\%+20\%) \\ \hline
    Tested on & PSNR/SSIM & PSNR/SSIM & PSNR/SSIM \\ \hline
    10\%      & \textbf{29.52}/\textbf{0.894} & 13.49/0.620 & \underline{17.81}/\underline{0.728} \\
    20\%      & 12.76/0.669 & \textbf{31.36}/\textbf{0.936} & \underline{31.17}/\underline{0.932} \\
    Mixed (10\%+20\%)  & 21.33/\underline{0.784} & \underline{22.24}/0.775 & \textbf{24.52}/\textbf{0.845} \\
    \hline
    \textit{Average} & 21.20/\underline{0.782} & \underline{22.36}/0.777 & \textbf{24.50}/\textbf{0.835} \\
    \bottomrule
    \end{tabular}
    \label{tab:studylightlevels}%\vspace{-3mm}
\end{table}

\begin{figure}[t!]
    \centering
    \includegraphics[trim={0mm 0cm  26.5cm 0mm}, clip, width=\columnwidth]{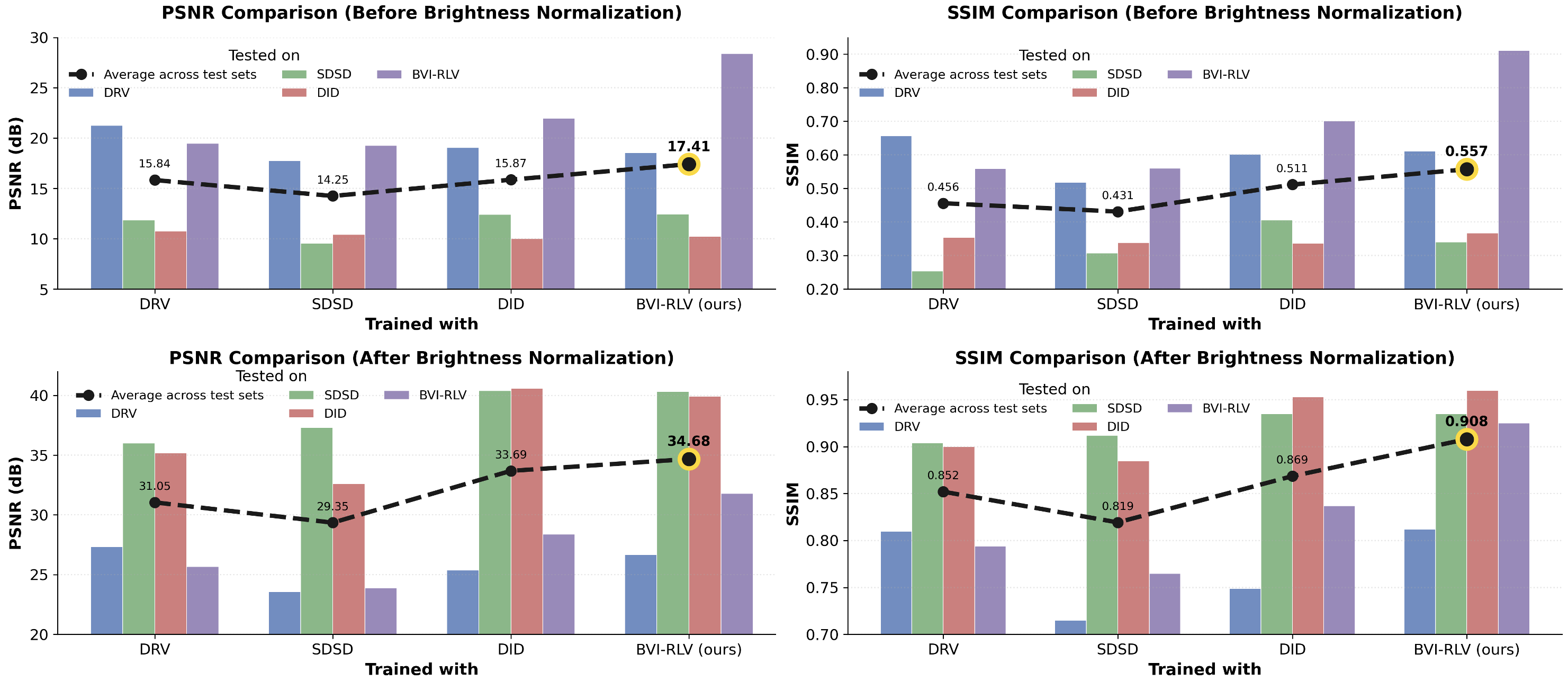} %\vspace{-3mm}
    \caption{Comparison of quantitative results of mo-CGAN across datasets before (top row) and after (bottom row) brightness normalization.}
    \label{fig:postprocess_comp} %\vspace{-3mm}
\end{figure}
% ------------------------------------------------------------
\subsubsection{Influence of different light levels}

BVI-RLV contains sequences captured at two controlled illumination levels (10\% and 20\% of maximum brightness), enabling analysis of lighting variability on robustness. A CNN-based model is trained under three settings: 10\%-only, 20\%-only, and mixed (10\%+20\%), and evaluated on both matched and cross-light conditions. As shown in Table~\ref{tab:studylightlevels}, performance is highest when training and testing illumination match, while the mixed-light model achieves the best overall robustness across all conditions.
Despite using only two illumination levels, models generalize well to unseen lighting. A preliminary study using linear brightness interpolation to simulate intermediate conditions shows no significant gain over the baseline, suggesting that two discrete illumination levels are sufficient for robust generalization.

These results highlight the importance of illumination diversity during training. Exposure to multiple lighting levels introduces a broader range of signal-to-noise ratios and degradation types, improving adaptability to unseen scenes. The mixed-light model improves cross-light performance by nearly +3.0~dB average PSNR over single-light training, confirming robustness to illumination shifts. In contrast, single-light models tend to overfit to a fixed exposure level (Fig.~\ref{fig:visualresults}, Fig.~\ref{fig:public_outdoor}), limiting generalization to real-world variations. This diversity in BVI-RLV directly contributes to its superior cross-dataset performance in Table~\ref{tab:alltest}.

% shortened version if necessary
% These results show that illumination diversity improves robustness. Multiple lighting levels expose the model to broader signal-to-noise ratios and degradation types, improving adaptation to unseen scenes. The mixed-light model improves average PSNR by nearly +3.0~dB over single-light training, confirming its robustness to illumination shifts and contributing to the cross-dataset gains in Table~\ref{tab:alltest}.

%Despite the training datasets containing only two brightness levels, the models generalize well to diverse illumination conditions, suggesting that explicitly incorporating additional brightness levels is unnecessary. This finding is further validated by our preliminary brightness augmentation ablation, in which we linearly interpolated between the two brightness levels during training, yet observed no significant improvement over the baseline without augmentation.

%--------------------------------------------------------

\subsubsection{Evaluating dataset quality after brightness normalization} 
While the previous experiment analyzed illumination variability within our dataset, this study examines the impact of cross-dataset brightness discrepancies on restoration evaluation, verifying that BVI-RLV’s gains are not due to brightness bias. We re-evaluate mo-CGAN using histogram-matching-based brightness normalization between enhanced sequences and reference frames to align illumination across datasets.
Fig.~\ref{fig:postprocess_comp} reports results before (top) and after (bottom) normalization. In both cases, BVI-RLV-trained models achieve the highest average PSNR, confirming that performance gains are consistent after removing illumination bias. This indicates that BVI-RLV improves true content restoration rather than benefiting from exposure bias. 

 \subsubsection{Failure case analysis} BVI-RLV-trained models show the smallest relative gain on DRV sequences that contain highly saturated artificial lighting (e.g.\ neon signs), whose spectral characteristics differ substantially from the 6500~K studio illumination of BVI-RLV. This represents a distribution shift that future extensions of the dataset could address through multi-illuminant capture.

\section{Conclusion}
This paper introduced BVI-RLV, a fully registered dataset for low-light video enhancement. It contains 40 dynamic scenes captured under two controlled low-light levels, each paired with normal-light ground truth and aligned with sub-pixel precision. Four baseline models were implemented for fair and reproducible benchmarking. Experiments show that BVI-RLV consistently outperforms existing datasets, delivering superior registration accuracy, motion modeling, and generalization. These results establish BVI-RLV as a benchmark dataset for robust, real-world low-light video enhancement.

% \section*{Acknowledgment}
% We would like to thank Dave Blackham from Esprit Film \& Television for his advice on video capturing.

%\clearpage
\bibliographystyle{IEEEtran}
\bibliography{main}

\end{document}

\title{Supplementary Material: BVI-RLV: A Fully Registered Dataset for Low-Light Video Enhancement}

\author{%
  Ruirui~Lin,  Guoxi~Huang,  {Joanne~Lin, Qi~Sun, Alexandra~Malyugina, David~R.~Bull,~\IEEEmembership{Fellow,~IEEE,} and Nantheera~Anantrasirichai,~\IEEEmembership{Member,~IEEE,}} 

  % <-this % stops a space
\thanks{All authors are with Visual Information Laboratory, Bristol Vision Institute (BVI), University of Bristol, Bristol, UK BS1 5DD. This work was supported by the UKRI MyWorld Strength in Places Programme (SIPF00006/1) and  EPSRC NIA (UKRI241) }}

\markboth{Journal of \LaTeX\ Class Files,~Vol.~14, No.~8, August~2021}%
{Shell \MakeLowercase{\textit{et al.}}: A Sample Article Using IEEEtran.cls for IEEE Journals}

\maketitle

\section{Dataset Failure case analysis}
Only approximately 0.76\% of frame pairs did not achieve sub-pixel alignment. Inspection of these cases reveals three primary scenarios: (i) abrupt nonlinear motion transitions, such as in Kitchen and Figures2, where dolly reversal introduces temporal lag beyond the search window; (ii) reflective surfaces, such as in Lego, Kitchen, and Wires, where specular highlights shift nonlinearly and can affect MAE-based frame selection; and (iii) large homogeneous regions, such as in Mario1 and Cloths2, where limited texture provides insufficient anchor points for reliable NCC-based verification. These scenarios could inform the design of future capture protocols or extended search strategies.

\section{Grouped Statistics}
To examine whether the observed gains are stable across different conditions, we conduct a fine-grained grouped analysis on the BVI-RLV test set. The same Mamba-based enhancement model is trained separately on DRV, SDSD, DID, and BVI-RLV, and evaluated under four grouping dimensions: motion speed, movement type, motion complexity, and illumination level. Motion speed was grouped using the mean optical-flow magnitude, $\mu_{\mathrm{flow}}$, between successive frames, estimated using RAFT~\cite{raft2020}. Here, $\mu_{\mathrm{flow}}$ denotes the frame-level mean magnitude of the estimated optical-flow vectors. The threshold of 2.41 pixels was set midway between the maximum value in the slow-motion subset and the minimum value in the fast-motion subset. The resulting slow- and fast-motion subsets have mean $\mu_{\mathrm{flow}}$ values of 1.90 and 6.17 pixels, respectively, with ranges of 1.67-2.14 and 2.67-20.21 pixels.

Table~\ref{tab:grouped_stats} reports per-subset PSNR, SSIM, and LPIPS. The model trained on BVI-RLV achieves the best performance across all subsets, i.e., across all 24 subgroup-metric comparisons. Its PSNR margins over the second-best training dataset range from +7.6 to +18.4 dB, indicating that the gains are not concentrated in a small subset of favorable cases. The advantage remains clear under challenging conditions, including fast motion (+11.7 dB over DID), rotational movement (+12.4 dB over DID), nonlinear motion (+14.1 dB over DRV), and the 20\% illumination subset (+18.4 dB over DID).

At the same time, the grouped results reveal remaining challenging cases. For the BVI-RLV-trained model, the 10\% illumination subset yields lower SSIM and higher LPIPS than the 20\% subset, showing that extreme low-light degradation remains difficult. The fast-motion subset also has lower PSNR than the slow-motion subset, as expected. The linear-motion subset yields the lower PSNR for the BVI-RLV-trained model, which likely reflects the specific content of this subset, such as sustained translation or motion blur. Nevertheless, BVI-RLV remains the best-performing training dataset in all these cases, suggesting consistently improved robustness across motion and illumination.

\begin{table*}[!t]
\centering
\caption{Fine-grained grouped analysis of the Mamba-based model trained on various datasets and evaluated on the BVI-RLV test set. Results are reported as PSNR (dB), SSIM, and LPIPS. $\mu_{\mathrm{flow}}$ denotes the average pixel displacement between successive frames. \textbf{Bold} and  \underline{underlined} indicate the best and second-best performance}
\label{tab:grouped_stats}
\renewcommand{\arraystretch}{1.15}
\setlength{\tabcolsep}{3.0pt}
\resizebox{\textwidth}{!}{
\begin{tabular}{@{}ll ccc ccc ccc ccc ccc@{}}
\toprule
\multirow{2}{*}{\textbf{Category}} 
& \multirow{2}{*}{\textbf{Subset}}
& \multicolumn{3}{c}{\textbf{DRV}~\cite{Chen_2019_ICCV}}
& \multicolumn{3}{c}{\textbf{SDSD}~\cite{wang2021sdsd}}
& \multicolumn{3}{c}{\textbf{DID}~\cite{Fu:dancing:2023}}
& \multicolumn{3}{c}{\textbf{BVI-RLV (Ours)}}\\
\cmidrule(lr){3-5} \cmidrule(lr){6-8} \cmidrule(lr){9-11} \cmidrule(lr){12-14}
& & PSNR & SSIM & LPIPS & PSNR & SSIM & LPIPS & PSNR & SSIM & LPIPS & PSNR & SSIM & LPIPS \\
\midrule

\multirow{2}{*}{Motion speed}
& Slow ($\mu_{\mathrm{flow}}\leq\,$2.41 pixels) 
& \underline{19.38} & 0.723 & \underline{0.239} & 18.78 & 0.726 & 0.287 & 18.46 & \underline{0.727} & 0.247 & \textbf{33.37} & \textbf{0.919} & \textbf{0.073} \\
& Fast ($\mu_{\mathrm{flow}}>\,$2.41 pixels) 
& 18.75 & 0.697 & \underline{0.242} & 17.16 & 0.692 & 0.316 & \underline{19.37} & \underline{0.727} & 0.256 & \textbf{31.08} & \textbf{0.910} & \textbf{0.074}\\
\midrule

\multirow{2}{*}{Movement type}
& Parallel 
& \underline{18.36} & 0.726 & \underline{0.235} & 17.25 & 0.710 &0.329 & 18.15 & \underline{0.735} & 0.283 & \textbf{31.79} & \textbf{0.917} & \textbf{0.071}  \\
& Rotational 
& 20.41 & 0.675 & 0.251 & 19.29 & 0.704 & 0.248 & \underline{20.54} & \underline{0.709} & \underline{0.220} & \textbf{32.89} & \textbf{0.909} & \textbf{0.077} \\
\midrule

\multirow{2}{*}{Motion complexity}
& Linear 
& 18.84 & 0.698 & \underline{0.250} & 17.85 & 0.700 & 0.316 & \underline{19.60} & \underline{0.725} & 0.284 & \textbf{29.75} & \textbf{0.907} & \textbf{0.077} \\
& Nonlinear 
& \underline{19.14} & 0.714 & \underline{0.236} & 17.97 & 0.711 & 0.295 & 18.67 & \underline{0.727} & 0.252 & \textbf{33.22} & \textbf{0.917} & \textbf{0.072} \\
\midrule

\multirow{2}{*}{Illumination level}
& 10\% 
& \underline{23.62} & 0.758 & \underline{0.246} & 21.15 & \underline{0.778} & 0.285 & 23.03 & 0.736 & 0.284 & \textbf{31.24} & \textbf{0.897} & \textbf{0.092} \\
& 20\% 
& 14.28 & 0.659 & \underline{0.234} & 14.58 & 0.636 & 0.319 & \underline{14.70} & \underline{0.718} & 0.238 &  \textbf{33.13} & \textbf{0.932} & \textbf{0.054} \\
\bottomrule
\end{tabular}
}
% \vspace{-2mm}
\end{table*}

\begin{figure*}[!t]
    \centering
    % \vspace{-1mm}
    \includegraphics[width=\linewidth]{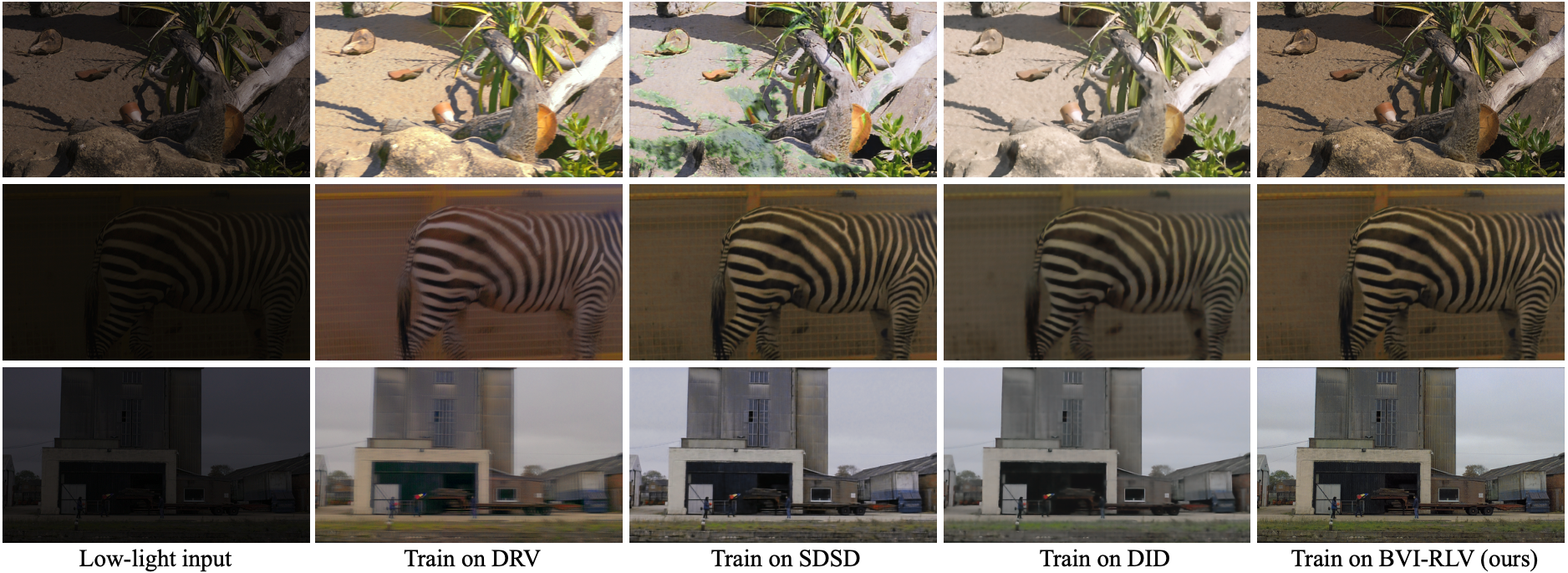}
    \caption{Qualitative results of the Mamba-based model trained on different datasets and tested on additional real-world outdoor low-light videos.}
    \label{fig:more_outdoor}
        % \vspace{-4mm}
\end{figure*}

\section{Additional Outdoor Visualization}
In addition to the outdoor examples shown in Section~V-B, we provide qualitative comparisons on additional \textit{real-world} outdoor low-light videos (no available ground truths), including wildlife and wide-area outdoor scenes, from a dataset described in a manuscript under review~\cite{lin2026olives}. The dataset is captured outdoors using professional cinema-grade, RED V-RAPTOR cameras at 1920$\times$1080 resolution and 25 fps.

As shown in Fig.~\ref{fig:more_outdoor}, the Mamba-based model trained on BVI-RLV achieves more balanced illumination and better preserves object details and scene structures, while models trained on existing datasets often introduce over-exposure, color shifts, blur, or local artifacts, especially in challenging outdoor scenes with complex motion and non-uniform illumination.

\section{Downstream Tasks}
Due to the lack of publicly available annotated benchmarks for real low-light videos, we use ELVIS-S~\cite{lin2026elvisenhancelowlightvideo} as an independent downstream evaluation set. ELVIS-S is a manually annotated \textit{real-world} low-light VIS dataset and is not used to train our enhancement models. As no paired normal-light videos are available, the ground-truth annotations are manual instance annotations on the original low-light frames. Since VIS jointly segments and associates instances across frames, it also reflects temporal tracking performance.

For downstream evaluations, we use a frozen MinVIS~\cite{huang2022minvis} model with a ResNet-50 backbone pretrained on YouTube-VIS 2019. This provides a standard VIS evaluator whose pretraining categories include ELVIS-S classes such as `monkey' and `sedan'. The selected Mamba-based enhancement method, which performs best in Table~3, is trained on each dataset, and its enhanced frames are evaluated using the same MinVIS model.

\noindent \textbf{Video instance segmentation.} As shown in Table~\ref{tab:vis_downstream}, BVI-RLV achieves the best AP, AP$_{50}$, AR$_1$, and AR$_{10}$, with a 5.28 AP gain over the second-best dataset, SDSD. Although DRV obtains the highest AP$_{75}$, BVI-RLV achieves stronger overall precision-recall performance and the highest recall. These results indicate that enhancement models trained on BVI-RLV better preserve instance-level cues under challenging low-light conditions, leading to stronger downstream performance for VIS.

\noindent \textbf{Object detection.} Table~\ref{tab:det_downstream} reports per-frame box-level detection results. Since MinVIS follows the Mask2Former~\cite{cheng2021mask2former} mask-classification paradigm, per-instance bounding boxes are obtained as the tight rectangles enclosing the predicted masks and are evaluated using the COCO protocol. Compared with mask AP, box AP is generally higher because bounding-box IoU is less sensitive to fine boundary errors than mask IoU. BVI-RLV again achieves the best AP, AP$_{50}$, and AR$_{10}$, improving over the second-best results by 3.40, 5.73, and 8.03 points, respectively. It also obtains the second-best AP$_{75}$ and AR$_1$, showing that the advantage of BVI-RLV generalizes from mask-level VIS to box-level detection.

\begin{figure*}[!ht]
    \centering
    % \vspace{-2mm}
\includegraphics[width=\linewidth]{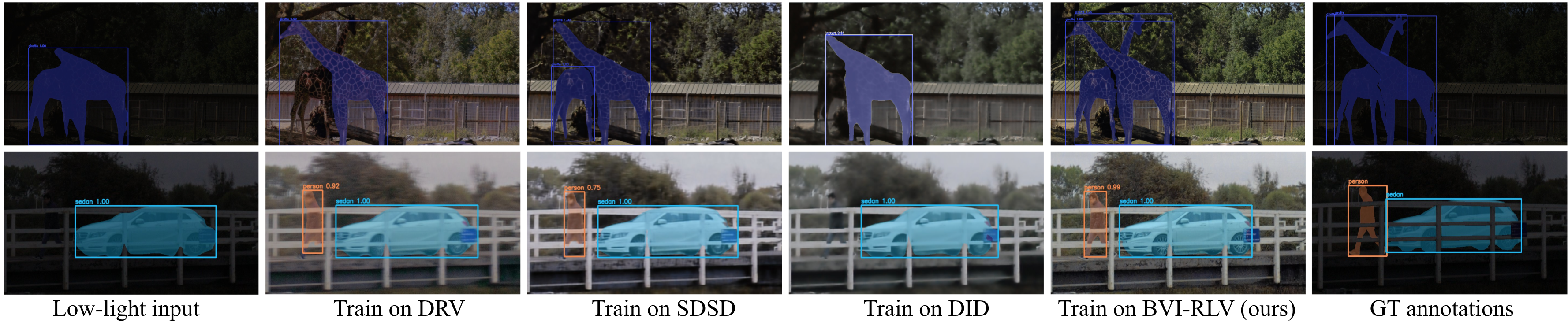}
    \caption{Qualitative downstream VIS results on ELVIS-S using a pretrained MinVIS model. Enhanced frames are generated by the Mamba-based model trained on different datasets. The last column shows ground-truth VIS annotations on the low-light frames.}
        % \vspace{-1mm}
    \label{fig:downstream_visual}
\end{figure*}

Qualitatively, Fig.~\ref{fig:downstream_visual} shows that the BVI-RLV-trained model produces cleaner instance masks and more complete detections, including clearer separation of the two giraffes and a high-confidence detection of the person adjacent to the sedan. Overall, the downstream results demonstrate that BVI-RLV is not only beneficial for low-light enhancement quality, but also improves the usability of enhanced videos for high-level perception tasks.

% Table 1: VIS evaluation (matches ELVIS)
\begin{table}[!t]
\centering
% \vspace{-1mm}
\caption{Downstream video instance segmentation evaluation on ELVIS-S~\cite{lin2026elvisenhancelowlightvideo} using a frozen MinVIS~\cite{huang2022minvis} with ResNet-50 backbone pretrained on YouTube-VIS 2019. \textbf{Bold} and \underline{underlined} indicate the best and second-best performance}
\label{tab:vis_downstream}
\renewcommand{\arraystretch}{1.15}
\setlength{\tabcolsep}{8pt}
\resizebox{\columnwidth}{!}{
\begin{tabular}{@{}l ccc cc@{}}
\toprule
\multirow{2}{*}{\textbf{Trained with}} 
& \multicolumn{3}{c}{\textbf{Mask AP}} 
& \multicolumn{2}{c}{\textbf{Mask AR}} \\
\cmidrule(lr){2-4} \cmidrule(lr){5-6}
& AP & AP$_{50}$ & AP$_{75}$ & AR$_{1}$ & AR$_{10}$ \\
\midrule
Low-light input & 32.50 & 50.00 & 25.00 & 32.50 & 32.50 \\
DRV~\cite{Chen_2019_ICCV} & 43.94 & 75.25 & \textbf{40.84} & \underline{42.19} & 43.75 \\
SDSD~\cite{wang2021sdsd} & \underline{46.70} & \underline{90.10} & 32.92 & 37.19 & \underline{50.62} \\
DID~\cite{Fu:dancing:2023} & 42.39 & 85.97 & 26.07 & 39.69 & 44.06 \\
BVI-RLV (Ours) & \textbf{51.98} & \textbf{96.78} & \underline{34.53} & \textbf{43.12} & \textbf{56.25} \\
\bottomrule
\end{tabular}}
% \vspace{-4mm}
\end{table}

% Table 2: per-frame object detection
\begin{table}[!t]
\centering
% \vspace{-1mm}
\caption{Downstream per-frame object detection on ELVIS-S~\cite{lin2026elvisenhancelowlightvideo}. The same pretrained MinVIS model and enhancement pipelines as Table~\ref{tab:vis_downstream} are used. \textbf{Bold} and \underline{underlined} indicate the best and second-best performance}
\label{tab:det_downstream}
\renewcommand{\arraystretch}{1.15}
\setlength{\tabcolsep}{8pt}
\resizebox{\columnwidth}{!}{
\begin{tabular}{@{}l ccc cc@{}}
\toprule
\multirow{2}{*}{\textbf{Trained with}} 
& \multicolumn{3}{c}{\textbf{Box AP}} 
& \multicolumn{2}{c}{\textbf{Box AR}} \\
\cmidrule(lr){2-4} \cmidrule(lr){5-6}
& AP & AP$_{50}$ & AP$_{75}$ & AR$_{1}$ & AR$_{10}$ \\
\midrule
Low-light input & 50.39 & 76.11 & 51.33 & 51.44 & 60.43 \\
DRV~\cite{Chen_2019_ICCV} & \underline{63.08} & 77.61 & \textbf{67.80} & \textbf{59.73} & 67.22 \\
SDSD~\cite{wang2021sdsd} & 59.47 & 76.00 & 66.48 & 51.34 & \underline{68.78} \\
DID~\cite{Fu:dancing:2023} & 58.02 & \underline{85.32} & 58.16 & 56.09 & 63.70 \\
BVI-RLV (Ours) & \textbf{66.48} & \textbf{91.05} & \underline{67.14} & \underline{57.80} & \textbf{76.81} \\
\bottomrule
\end{tabular}}
% \vspace{-4mm}
\end{table}

\newpage
\bibliographystyle{IEEEtran}
\bibliography{main}

\newpage

\vfill